\documentclass[final,5p,times,twocolumn,authoryear]{elsarticle}

\usepackage{amssymb}
\usepackage{amsthm}
\usepackage{amsmath}
\usepackage{booktabs}
\usepackage{diagbox}
\usepackage{enumerate}
\usepackage{color,xcolor}

\newtheorem{thm}{Theorem}[section]
\newtheorem{exa}[thm]{Example}
\newtheorem{pro}[thm]{Proposition}
\newtheorem{defi}[thm]{Definition}
\newtheorem{lemma}[thm]{Lemma}

\newtheorem{rem}[thm]{Remark}

\newcommand{\R}{\mathbb{R}}
\newcommand{\T}{\mathcal{T}}
\newcommand{\E}{\mathcal{E}}
\newcommand{\rnum}[1]{\romannumeral #1}%
\newcommand{\RNum}[1]{\uppercase\expandafter{\romannumeral #1\relax}}%

\journal{ }

\begin{document}

\begin{frontmatter}



\title{Quantifying the generalization error in deep learning in terms of data distribution and neural network smoothness}


\author[cas,ucas]{Pengzhan Jin\fnref{cof}}
\author[brown]{Lu Lu\fnref{cof}}
\author[cas,ucas]{Yifa Tang}
\author[brown]{George Em Karniadakis\corref{cor}}
\cortext[cor]{Corresponding author at: 182 George Street, Providence, RI 02912, USA.}
\ead{george\_karniadakis@brown.edu}
\fntext[cof]{Pengzhan Jin and Lu Lu contributed equally to this work.}

\address[cas]{LSEC, ICMSEC, Academy of Mathematics and Systems Science, Chinese Academy of Sciences, Beijing 100190, China}
\address[ucas]{School of Mathematical Sciences, University of Chinese Academy of Sciences, Beijing 100049, China}
\address[brown]{Division of Applied Mathematics, Brown University, Providence, RI 02912, USA}

\begin{abstract}
The accuracy of deep learning, i.e., deep neural networks, can be characterized by dividing the total error into three main types:  approximation error, optimization error, and generalization error. Whereas there are some satisfactory answers to the problems of approximation and optimization, much less is known about the theory of generalization. Most existing theoretical works for generalization fail to explain the performance of neural networks in practice. To derive a meaningful bound, we study the generalization error of neural networks for classification problems in terms of data distribution and neural network smoothness. We introduce the \textit{cover complexity} (CC) to measure the difficulty of learning a data set and the \textit{inverse of the modulus of continuity} to quantify neural network smoothness. A quantitative bound for expected accuracy/error is derived by considering both the CC and neural network smoothness. Although most of the analysis is general and not specific to neural networks, we validate our theoretical assumptions and results numerically for neural networks by several data sets of images. The numerical results confirm that the expected error of trained networks scaled with the square root of the number of classes has a linear relationship with respect to the CC. We also observe a clear consistency between test loss and neural network smoothness during the training process. In addition, we demonstrate empirically that the neural network smoothness decreases when the network size increases whereas the smoothness is insensitive to training dataset size.
\end{abstract}




\end{frontmatter}


\section{Introduction}
\label{intro}

\begin{figure}[htpb]
    \centering
    \includegraphics[width=0.48\textwidth]{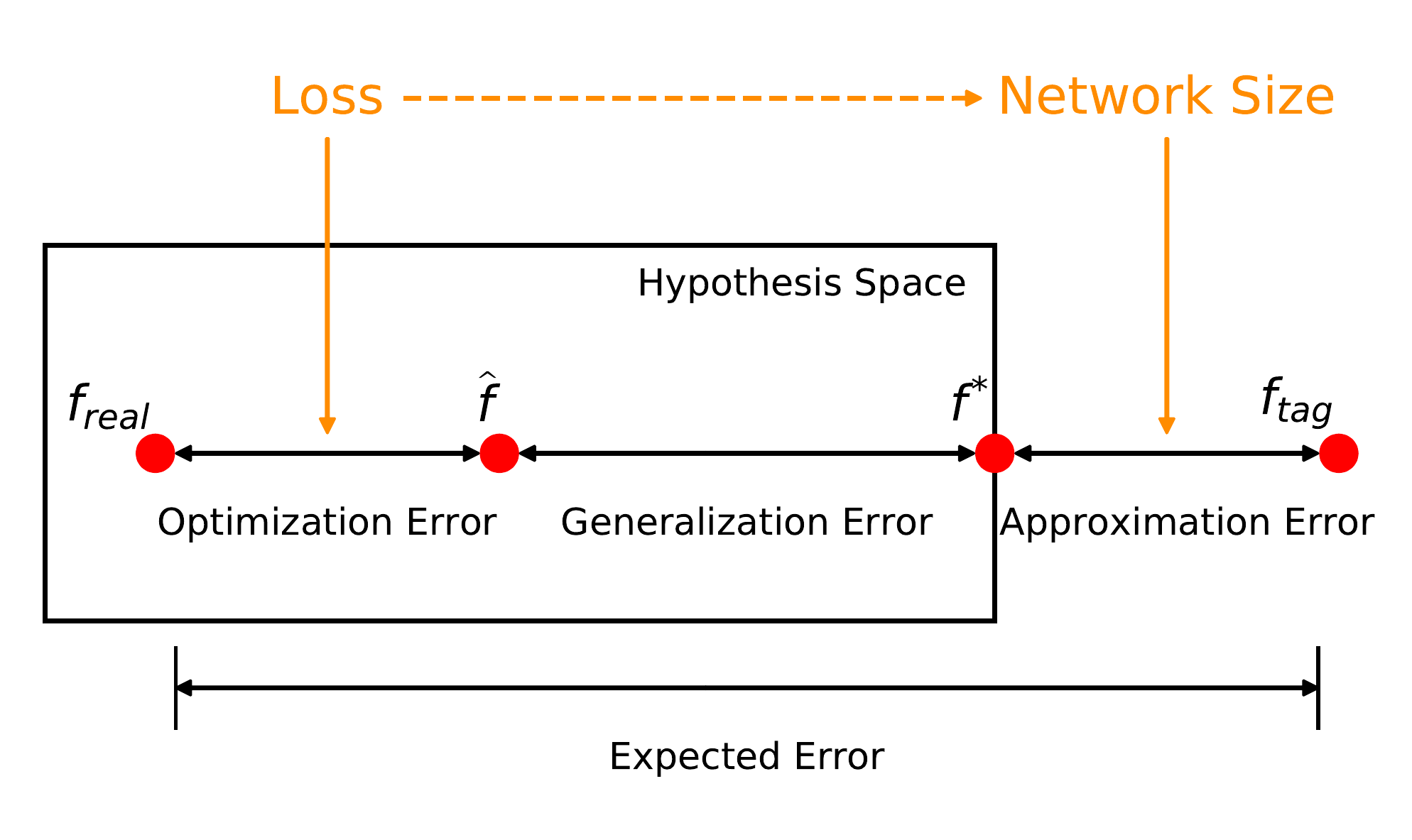}
    \caption{Illustration of approximation error, optimization error, and generalization error. The total error consists of these three errors. $f_{tag}$ is the target ground-truth function, $f^{*}$ is the function closest to $f_{tag}$ in the hypothesis space, $\hat{f}$ is a neural network whose loss is at a global minimum of the empirical loss, and $f_{real}$ is the function returned by the training algorithm. Thus, the optimization error is correlated with the value of the empirical loss, while the approximation error depends on the network size. In addition, a small loss requires a large network size, which in turn leads to a small approximation error. Assuming a sufficiently small empirical loss, the expected error mainly depends on the generalization error.}
    \label{fig:error_types}
\end{figure}

In the last 15 years, deep learning, i.e., deep neural networks (NNs), has been used very effectively in diverse applications, such as image classification~\citep{krizhevsky2012imagenet}, natural language processing~\citep{maas2013rectifier}, and game playing~\citep{silver2016mastering}. Despite this remarkable success, our theoretical understanding of deep learning is lagging behind. The accuracy of NNs can be characterized by dividing the expected error into three main types: approximation (also called expressivity), optimization, and generalization~\citep{bottou2008tradeoffs,bottou2010large}, see Fig.~\ref{fig:error_types}. The well-known universal approximation theorem was obtained by~\citet{cybenko1989approximation} and~\citet{hornik1989multilayer} almost three decades ago stating that feed-forward neural nets can approximate essentially any function if their size is sufficiently large. In the past several years, there have been numerous studies that analyze the landscape of the non-convex objective functions, and the optimization process by stochastic gradient descent (SGD)~\citep{lee2016gradient,liao2017theory,lu2018collapse,allen2018convergence,du2018gradient,lu2019dying}. Whereas there are some satisfactory answers to the problems of approximation and optimization, much less is known about the theory of generalization, which is the focus of this study.

The classical analysis of generalization is based on controlling the complexity of the function class, i.e., model complexity, by managing the bias-variance trade-off~\citep{friedman2001elements}. However, this type of analysis is not able to explain the small generalization gap between training and test performance of neural networks learned by SGD in practice, considering the fact that deep neural networks often have far more model parameters than the number of samples they are trained on, and have sufficient capacity to memorize random labels~\citep{neyshabur2014search,zhang2016understanding}. To explain this phenomenon, several approaches have been recently developed by many researchers. The first approach is characterizing neural networks with some other low ``complexity'' instead of the traditional Vapnik-Chervonenkis (VC) dimension~\citep{bartlett2017nearly} or Rademacher complexity~\citep{bartlett2002rademacher}, such as path-norm~\citep{neyshabur2015path}, margin-based bounds~\citep{sokolic2017robust,bartlett2017spectrally,neyshabur2017pac}, Fisher-Rao norm~\citep{liang2017fisher}, and more~\citep{neyshabur2018role,wei2019data}. The second approach is to analyze some good properties of SGD or its variants, including its stability~\citep{hardt2015train,kuzborskij2017data,gonen2017fast,chen2018stability}, robustness~\citep{sokolic2016generalization,sokolic2017robust}, implicit biases/regularization~\citep{poggio2017theoryiii,soudry2018implicit,gunasekar2018implicit,nagarajan2019generalization}, and the structural properties (e.g., sharpness) of the obtained minimizers~\citep{keskar2016large,dinh2017sharp,zhang2018theory}. The third approach relies on overparameterization, e.g., sufficiently overparameterized networks can learn the ground truth with a small generalization error using SGD from random initialization~\citep{li2018learning,allen2018learning,arora2019fine,cao2019generalization}. There are also other approaches, such as compression~\citep{arora2018stronger,baykal2018data,zhou2018compressibility,cheng2018model}, Fourier analysis~\citep{rahaman2018spectral,xu2019frequency}, ``double descent'' risk curve~\citep{belkin2018reconciling}, PAC-Bayesian framework~\citep{neyshabur2017pac,nagarajan2018deterministic}, and information bottleneck~\citep{shwartz2017opening,saxe2019information}.

However, most theoretical bounds fail to explain the performance of neural networks in practice~\citep{neyshabur2017exploring,arora2018stronger}. To get non-vacuous and tight enough bounds to be practically meaningful, some problem-specific factors should be taken into consideration, such as the low complexity (i.e., data-dependent analysis)~\citep{dziugaite2017computing,kawaguchi2017generalization}, or properties of the trained neural networks~\citep{sokolic2017robust,arora2018stronger,wei2019data}. In this study, to achieve a practically meaningful bound, our analysis relies on the data distribution and the smoothness of the trained neural network. The analysis proposed in this study provides guarantees on the generalization error, and theoretical insights to guide the practical application.

\begin{figure*}[htbp]
    \centering
    \includegraphics[width=\textwidth]{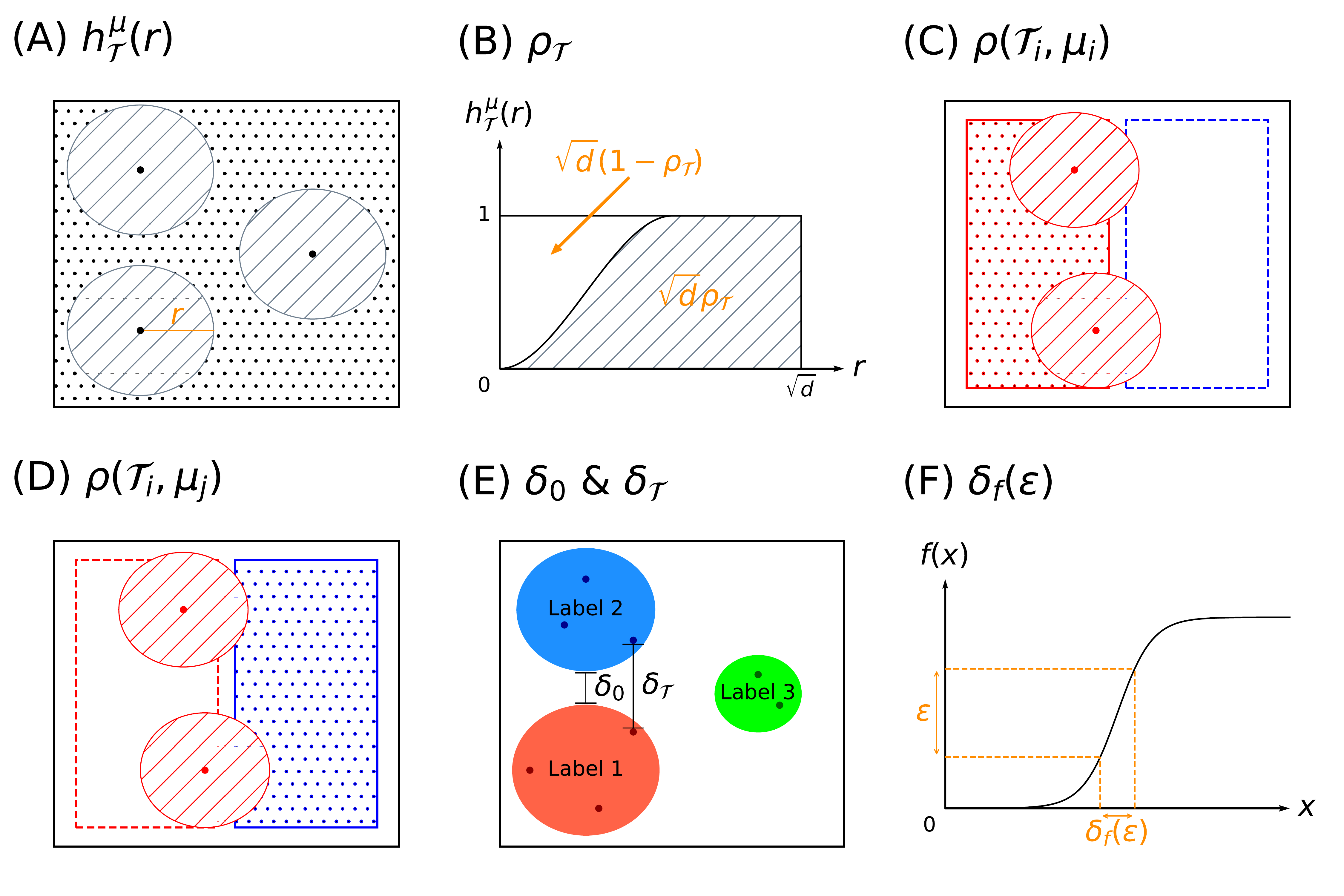}
    \caption{Illustrations of the main definitions and terminologies. (\textbf{A}) $h_{\T}^{\mu}(r)$ in Eq. (\ref{eq:h_T_r}): the probability of the neighborhood of the training set with radius of $r$. (\textbf{B}) $\rho_{\T}$ in Definition \ref{def:covers}: total cover of training set $\T$. (\textbf{C}) $\rho(\T_{i},\mu_{i})$ in Definition \ref{def:covers}: self cover of training set $\T$, which intuitively represents the aggregation of the data points of the same class. The red dots denote the probability distribution of the class ``red'', and the area of the disks intersected with the dotted rectangle represents the probability of the neighbourhood of two ``red'' data points with respect to the distribution of the class``red'' itself. (\textbf{D}) $\rho(\T_{i},\mu_{j})$ in Definition \ref{def:covers}: mutual cover of training set $\T$, which intuitively represents the overlapping between the data points of two different classes. The blue dots denote the probability distribution of the class ``blue'', and the area of the disks intersected with the dotted rectangle represents the probability of the neighbourhood of the two ``red'' data points with respect to the distribution of the class ``blue''. (\textbf{E}) $\delta_0$ in Proposition~\ref{pro:separation_gap}: separation gap; $\delta_{\T}$ in Theorem \ref{thm:deltaf_bound}: empirical separation gap. (\textbf{F}) $\delta_{f}$ in Definition \ref{def:delta_f}: the inverse of the modulus of continuity.}
    \label{fig:notation_illustration}
\end{figure*}

As shown in Fig.~\ref{fig:error_types}, the optimization error is correlated with the loss value (for notation simplicity, the term ``loss" indicates ``empirical loss"), while the approximation error depends on the network size. In addition, a small loss requires a sufficient approximation ability, i.e., a large network size, which in turn leads to a small approximation error. If we assume a sufficiently small loss, which usually holds in practice, then the expected error mainly depends on the generalization error. Hence, we study the expected error/accuracy directly. In particular, we propose a mathematical framework to analyze the expected accuracy of neural networks for classification problems. We introduce the concepts of \textit{total cover (TC)}, \textit{self cover (SC)}, \textit{mutual cover (MC)} and \textit{cover difference (CD)} to represent the data distribution, and then we use the concept of \textit{cover complexity (CC)} as a measure of the complexity of classification problems. On the other hand, the smoothness of a neural network $f$ is characterized by the \textit{inverse of the modulus of continuity} $\delta_f$. Because computing $\delta_f$ is not tractable in general, we propose an estimation using the spectral norm of the weight matrices of the neural network. The main terminologies are illustrated in Fig~\ref{fig:notation_illustration}. By combining the properties of the data distribution and the smoothness of neural networks, we derive a lower bound for the expected accuracy, i.e., an upper bound for the expected classification error.

Subsequently, we test our theoretical bounds on several data sets, including MNIST~\citep{lecun1998gradient}, CIFAR-10~\citep{krizhevsky2009learning}, CIFAR-100~\citep{krizhevsky2009learning}, COIL-20~\citep{nene1996columbia}, COIL-100~\citep{nene1996object}, and SVHN~\citep{netzer2011reading}. Our numerical results not only confirm our theoretical bounds, but also provide insights into the optimization process and the learnability of neural networks. In particular, we find that:
\begin{itemize}
    \item The best accuracy that can be achieved in practice (i.e., optimized by stochastic gradient descent) by fully-connected networks is approximately linear with respect to the cover complexity of the data set.
    \item The trend of the expected accuracy is consistent with the smoothness of the neural network, which provides a new ``early stopping'' strategy by monitoring the smoothness of the neural network.
    \item The neural network smoothness decreases when the network depth and width increases, with the effects of depth more significant than that of width.
    \item The neural network smoothness is insensitive to the training dataset size, and is bounded from below by a positive constant. This point makes our theoretical result (Theorem~\ref{thm:bounds}) specifically pertinent to deep neural networks.
\end{itemize}

The paper is organized as follows. After setting up notation and terminology in Section \ref{prel}, we present the main theoretical bounds for the accuracy based on the data distribution and the smoothness of neural networks in Section \ref{accu}, while all proofs are deferred to the appendix. In Section \ref{num_res}, we provide the numerical results for several data sets. In Section \ref{disc} we include a discussion, and in Section \ref{conc} we summarize our findings.

\section{Preliminaries} \label{prel}

Before giving the main results, we introduce the necessary notation and terminology. Without loss of generality, we assume that the space we need to classify is
$$D=[0,1]^{d},$$
where $d$ is the dimensionality, and the points in this space are classified into $K$ categories, i.e., there are $K$ labels $\{1,2,\cdots,K\}$. We denote the probability measure on $D$ by $\mu$, i.e., for a measurable set $A\subseteq D$, $\mu(A)$ is the probability of a random sample belonging to $A$.

\subsection{Ideal label function}

For the problem setup, we assume that every sample has at least one true label, and one sample may have multiple true labels. Taking image classification as an example, each image has at least one correct label. A fuzzy image or an image with more than one object in it may have multiple possible correct labels, and as long as the prediction is one of these labels, we consider the prediction to be correct.

It is intuitive that when two samples are close enough, they should have similar labels, which means that the ideal label function should be continuous. The continuity of a mapping depends on the topology of both domain and image space. For the domain of the ideal label function, we choose the standard topology induced by the Euclidean metric. As for the topology of the image space, we define it as follows. We first define the label set and the topology on it.

\begin{defi}[Topology] \label{def:topo}
Let
$$T=2^{\{1,2,\cdots,K\}}\backslash \{\varnothing\}$$
be the label set. Define the topology on $T$ to be
$$\tau_{T}=\left\{\bigcup_{k\in I}U_{k}\Bigg|I\subseteq T\right\},$$
where
$U_{k}=\{V\in T|V\supseteq k\}$ for $k\in T$, and thus $(T,\tau_{T})$ constitutes a topological space.
\end{defi}
Analogous to the Euclidean metric topology, $U_{k}$ is viewed as the open ``ball'' centered at $k$, and arbitrary unions of the ``balls'' $U_{k}$ are defined as the open sets, see \ref{exam:topo} for an example. With this choice of the topology, a function $f:D\to T$ is continuous if and only if
$$\forall x\in D\forall k\in f(x)\exists \delta>0\ s.t.\ k\in f(y)\ if\ \|y-x\|<\delta.$$
Next we give the definition of the ideal label function according to this topological space.

\begin{defi}[Ideal label function]
An idea label function is a continuous function $$tag:D\to T$$ where $D$ is equipped with the Euclidean metric topology and $T$ with the topology $\tau_{T}$ from Definition \ref{def:topo}. This continuity holds if and only if
\begin{equation}  \label{eq2}
\forall x\in D\forall k \in tag(x)\exists\delta>0\ s.t.\  k\in tag(y)\ if\  \|y-x\|<\delta.
\end{equation}
\end{defi}

Eq. (\ref{eq2}) means that two neighboring points would have some common labels. Based on the topological space defined above, it is easy to show that Eq. (\ref{eq2}) is equivalent to continuity. The reason why we consider a multi-label setup for classification problems is that it allows for the continuity property in Eq. (\ref{eq2}), which is impossible in the setup of a single label set, unless the label function is constant. In addition, the multi-label setup introduces a smooth transition, i.e., a buffer domain, between two domains of different labels, while the transition is sharp in the single label setup. In the following proposition, we show that if two samples are close enough, they must share at least one common label.

\begin{pro}[Separation gap] \label{pro:separation_gap}
$\exists\delta>0,\ s.t.\ tag(x)\cap tag(y)\neq \varnothing\ when\ \|x-y\|<\delta.$ We denote the supremum of $\delta$ as the separation gap $\delta_0$, which is used in the sequel.
\end{pro}
\begin{proof}
The proof can be found in ~\ref{proof1}.
\end{proof}

To understand the geometric interpretation of $\delta_{0}$, we consider the following special case: the label of each sample is either a single label set, such as $\{1\}$, or the full label set $\{1,2,\cdots,K\}$ if it is not uniquely identifiable.

\begin{pro}[Geometric interpretation of separation gap] \label{pro:geo_interpretation}
If the label of each sample is either a single label set or the full label set $\{1,2,\cdots,K\}$, then $\delta_{0}$ is the smallest distance between two different single label points, i.e.,
\begin{equation*}
\begin{split}
\delta_{0} = \inf \{\ &\|x-y\|\ |\ tag(x)\neq tag(y), and \\ &tag(x),tag(y)\in\{\{1\},\{2\},\cdots,\{K\}\}\ \}.
\end{split}
\end{equation*}
\end{pro}
\begin{proof}
The proof can be found in ~\ref{proof:geo_interpretation}.
\end{proof}

\subsection{Cover complexity of data set}

In this subsection, we introduce a quantity to measure the difficulty of learning a training data set 
$$\T=\{x_{1},x_{2},\dots,x_{n}\}\subseteq D.$$
First, we give some notations and propositions.  

With the measure $\mu$, the probability of the neighborhood of the training set with radius of $r$ is defined as
\begin{equation} \label{eq:h_T_r}
h_{\T}^{\mu}(r) := \mu\left(D\cap \bigcup\limits_{x_{i}\in \T}B(x_{i},r)\right),    
\end{equation}
where $B(x_{i},r)$ is the open ball centered at $x_{i}$ with radius of $r$, see Fig.~\ref{fig:notation_illustration}A. Obviously, $h_{\T}^{\mu}(r)$ is a monotone non-decreasing function, $h_{\T}^{\mu}(0)=0$ (since $B(x,0)=\varnothing$), and $h_{\T}^{\mu}(r)=1$ when $r> \sqrt{d}$, see Fig.~\ref{fig:notation_illustration}B. To represent the global behavior of $h_{\T}^{\mu}(r)$, we use the integral of $h_{\T}^{\mu}(r)$ with respect to $r$:
\begin{equation*}
\rho(\T,\mu) := \frac{1}{\sqrt{d}}\int_{0}^{\sqrt{d}}h_{\T}^{\mu}(r)dr.
\end{equation*}
Hence, $\rho(\T,\mu)$ considers both the number and location of the data points, and also the probability distribution of the space. The value $\rho(\T,\mu)$ is larger if the number of data points is increased and also if the probability distribution is more concentrated around $\T$, which we call the ``coverability'' of $\T$. We can increase $\rho(\T,\mu)$ by adding more data points or redistribute their locations. Next, we introduce the formal definition for the ``coverability''.

\begin{defi}[Coverability] \label{def:covers}
Let $\T$ be a data set from a domain $D$ with probability measure $\mu$. We define the following for the coverability of $\T$.
\begin{enumerate}[(i)]
\item The total cover (TC) is
  $$\rho_{\T} := \rho(\T,\mu).$$
  Thus, $0\leq\rho_{\T}\leq 1$.
\item The cover difference (CD) is
  $$CD(\T) := \frac{1}{K}\sum_{i}\rho(\T_{i},\mu_{i})-\frac{1}{K(K-1)}\sum_{i\neq j}\rho(\T_{i},\mu_{j}),$$
  where $K$ is the number of categories, and $\T_{i} \subset \T$ and $\mu_{i}$ represent the subset and probability measure of the label $i$ respectively, i.e.,
  \begin{equation*}
      \T_{i} := \{x\in\T|i\in tag(x)\},\quad \mu_{i}(A) := \frac{\mu(A\cap D_{i})}{\mu(D_{i})}
  \end{equation*}
  with $D_{i} := \{x\in D|i\in tag(x)\}$. Here, $\rho(\T_{i},\mu_{i})$ is called self cover (SC), and $\rho(\T_{i},\mu_{j})$ is called mutual cover (MC).
\item The cover complexity (CC) is
  $$CC(\T) := \frac{1-\rho_{\T}}{CD(\T)}$$
  for $CD(\T)\neq 0$.
\end{enumerate}
\end{defi}
\begin{rem}
CD is defined as the difference between the mean of SC and the mean of MC, since each category occurs with the same probability ($\sim 1/K$) in the data sets mostly used in practice. If there are some categories occurring more frequently than others, then it is straightforward to extend this definition by using the mean weighted by the probability of each category.
\end{rem}
In image classification, the dimension of the image space is very high, and thus the data points are quite sparse. However, due to the fact that images actually live on a manifold of low dimension, the probability density around $\T$ is actually high, which makes the TC to be meaningful. In our next result, we derive a lower bound of $h_{\T}^{\mu}(r)$ by $\rho_{\T}$.

\begin{pro} \label{pro:h_estimation}
Let $\T$ be a data set. $h_{\T}^{\mu}(r)$ and $\rho_{\T}$ are defined as above. Then we have

\begin{equation*}
h_{\T}^{\mu}(r)\geq 1-\frac{\sqrt{d}}{r}(1-\rho_{\T}),\quad 0<r\leq\sqrt{d}.
\end{equation*}

\end{pro}
\begin{proof}
The proof can be found in ~\ref{proof3}.
\end{proof}
From this proposition, we know that for a fixed $r$, $h_{\T}^{\mu}(r)$ is close to 1 if $\rho_{\T}$ is large enough. However, the probability distribution is usually given in practice, and we can only control the number of samples. The following theorem shows that $\rho_{\T}$ can be arbitrary close to 1 when enough samples are available.

\begin{thm}
Let $\T$ be a data set of size $n$ drawn according to $\mu$. Then there exists a non-increasing function $\varrho(\epsilon)$ satisfying $\lim\limits_{\epsilon\to0}\varrho(\epsilon)=1$, and for any $0<\eta,\epsilon\leq\frac{1}{2}$, there exists an
\begin{equation*}
m=O\left(\frac{d+1}{\epsilon}\ln\frac{d+1}{\epsilon}+\frac{1}{\epsilon}\ln\frac{1}{\eta}\right),
\end{equation*}
such that
\begin{equation*}
\rho_{\T} \geq \varrho(\epsilon)
\end{equation*}
holds with probability at least $1-\eta$ when $n\geq m$.
\end{thm}
\begin{proof}
The proof and some other results regarding TC can be found in~\ref{ETC}.
\end{proof} 

The reason why CD is introduced is that TC does not consider the labels of the data points. However, data points of the same label should be clustered in an easily learnable data set. $CD(\T)$ is the difference of self cover and mutual cover, which considers the distributions of each label. By normalizing TC with CD, the cover complexity $CC(\T)$ is able to measure the difficulty of learning a data set. The difficulty of a problem should be translation-independent and scale-independent. It is easy to see that $CC(\T)$ is independence of translation, and the following proposition shows that it is also scale-independent.

\begin{pro}[Scale independence] \label{pro:scale_independent}
$CC(\T)$ is scale-independent, i.e., if all the data points are scaled by the same factor less than 1, then $CC(\T)$ is unchanged.
\end{pro}
\begin{proof}
The proof can be found in ~\ref{proof:scale_independent}.
\end{proof}

\subsection{Setup for accuracy analysis}

The setup for accuracy analysis is as follows.
\begin{defi}
If $f:D\rightarrow \mathbb{R}^{K}$ is a continuous mapping, then the mapping
$$\hat{f}: x \mapsto \frac{e^{f(x)}}{\sum_i e^{f_{i}(x)}}$$
is still continuous, where $f_{i}(x)$ represents the i-th component of $f(x)$, and $e^{f(x)}\in \R^{K}$ is the exponential function applied componentwise to $f(x)$. We have $\hat{f}_i (x) > 0$, and $\sum_i \hat{f}_{i}(x)=1$. For convenience, we directly consider the case that $f_{i}(x)>0$ and $\sum f_{i}(x)=1$, and we call such mapping the normalized continuous positive mapping.
\end{defi}
\begin{rem}
A neural network with softmax nonlinearity is a normalized continuous positive mapping.
\end{rem}

Different from the accuracy usually used in classification problems, we define a stronger accuracy called $c$-accuracy as follows.
\begin{defi}[$c$-accuracy at $x$]
Let $f$ be a normalized continuous positive mapping. For $0.5\leq c<1$, we say that $f$ is $c$-accurate at point $x$ if
$$\exists i_{max}\in tag(x),\ s.t.\ f_{i_{max}}(x)>c.$$
\end{defi}

\begin{defi}[$c$-accuracy on $D$]
Let $f$ be a normalized continuous positive mapping. The $c$-accuracy of $f$ on a sample space $D$ is defined as
$$p_{c}(f) := \frac{\mu(H_{c}^{f})}{\mu(D)}=\mu(H_{c}^{f}),$$
where $H_{c}^{f}:=\{x\in D|f$ is $c$-accurate at $x$$\}$.
\end{defi}

\begin{defi}[$c$-accuracy on $\T$]
Let $f$ be a normalized continuous positive mapping. The $c$-accuracy of $f$ on a data set $\T$ is defined as
$$p_{c}^{\T} := \frac{\rho_{\widetilde{\T}_c}}{\rho_{\T}},$$
where $\widetilde{\T}_c:=\{x\in \T|f$ is $c$-accurate at $x\}$, and $\rho_{\widetilde{\T}_c}$ and $\rho_{\T}$ are the TC of $\widetilde{\T}_c$ and $\T$, respectively.
\end{defi}

\begin{defi}[Expected accuracy]
Let $f$ be a normalized continuous positive mapping. The expected accuracy of $f$ on a sample space $D$ is defined as
$$p(f) := \frac{\mu(H^{f})}{\mu(D)}=\mu(H^{f}),$$
where $H^{f} := \{x\in D|\exists i_{max}\in tag(x),\ s.t.\ f_{i_{max}}(x)> f_{i}(x)\ \forall 1\leq i\leq K,\ i\neq i_{max}\}$.
\end{defi}

We note that the $c$-accuracy of $f$ on $D$ represents the expected $c$-accuracy, and the $c$-accuracy of $f$ on $\T$ represents the empirical $c$-accuracy.

Finally, we define a non-decreasing function $\delta_{f}$ to describe the smoothness of $f$.

\begin{defi}[Smoothness] \label{def:delta_f}
Let $f:D \to \R^{K}$ be a continuous mapping. Then $f$ is uniformly continuous due to the compactness of $D$, i.e.
\begin{equation*}
\forall\epsilon >0,\ \exists \delta>0,\ s.t.\ \|f(x)-f(y)\|_{\infty}<\epsilon\ when\ \|x-y\|<\delta.
\end{equation*}
We denote the supremum of $\delta$ satisfying the above requirement by $\delta_{f}(\epsilon)$. It is easy to see that $\delta_{f}(\epsilon)$ is equal to the inverse of the modulus of continuity of $f$.
\end{defi}

For low dimensional problems, we can directly compute $\delta_{f}$ by brute force. However, for high dimensional problems, it is intractable to compute $\delta_{f}$, and thus we give the following lower bound of $\delta_{f}$ for a fully-connected ReLU-network $f$ with softmax as the activation function in the last layer, which is also the main network structure considered through this work. 

A fully-connected neural network is defined as follows:
$$\phi_{i}(x)=\sigma(W_{i}x+b_{i}),1\leq i\leq l-1,$$
$$f(x)=softmax(W_{l}(\phi_{l-1}\circ\cdots\circ\phi_{1}(x))+b_{l}),$$
$$W_{i}\in\R^{n_{i}\times n_{i-1}},b_{i}\in\R^{n_{i}},1\leq i\leq l,$$
where $n_{i}$ is the number of neurons in the layer $i$ ($n_{0}=d$ and $n_{l}=K$), and $\sigma$ is the activation function. Then for the ReLU activation function, we have
\begin{equation}\label{eq:lip}
\delta_{f}(\epsilon)\geq\frac{\epsilon}{Lip(f)}\geq\frac{\epsilon}{\|W_{1}\|_{2}\cdots\|W_{l}\|_{2}\cdot Lip(softmax)},
\end{equation}
where $\|\cdot\|_{2}$ is the spectral norm, $Lip(f)$ and $Lip(softmax)$ represent the Lipschitz constants of $f$ and $softmax$ mapping from $(D,\|\cdot\|_2)$ to $(\R^{K},\|\cdot\|_\infty)$, respectively. $Lip(softmax)$ is a constant less than $1/2$, and thus is ignored in our numerical examples. We note that although the lower bound of $\delta_f(\epsilon)$ depends exponentially on the neural net depth, $\delta_f(\epsilon)$ itself does not necessarily scale exponentially with the network depth.

\section{Lower bounds for the expected accuracy} \label{accu}

In this section, we present a theoretical analysis of the lower bound for the expected accuracy as well as an upper bound for the expected error.

\begin{pro} \label{pro:empirical_bound}
Let $f$ be a normalized continuous positive mapping. Suppose that $\T$ is a single label training set, i.e. $tag(\T)\subseteq \{\{1\},\{2\},\cdots,\{K\}\}$. For any $0.5\leq c_{2}<c_{1}\leq 1$, we have
\begin{equation*}
p_{c_{2}}(f)\geq 1-\frac{\sqrt{d}}{\delta}(1-p_{c_{1}}^{\T}\rho_{\T}),
\end{equation*}
where $\delta=\min(\delta_{0},\delta_{f}(c_{1}-c_{2}))$.
\end{pro}
\begin{proof}
The proof can be found in ~\ref{proof:empirical_bound}.
\end{proof}

Proposition \ref{pro:empirical_bound} shows that the expected $c_{2}$-accuracy of $f$ can be bounded by the empirical $c_{1}$-accuracy and the TC of the training set. We can see that $p_{c_{2}}(f)$ tends to 1 when $\rho_{\T}$ and $p_{c_{1}}^{\T}$ tend to 1. Next we derive a bound for the accuracy by taking into account the loss function.

\begin{thm}[Lower bound of $c$-accuracy]  \label{thm:lower_bound}
Let $f$ be a normalized continuous positive mapping. Suppose that $\T=\{x_{1},\cdots,x_{n}\}$ is a single label training set, and $tag(x_{i})=\{k_{i}\}$. For any $0.5\leq c<1$, if the maximum cross entropy loss
$$L_{f}^{max}:=\max\limits_{1\leq i\leq n}\ell (f(x_{i}),k_{i})=-\min\limits_{1\leq i\leq n}\ln(f_{k_{i}}(x_{i})) <-\ln c,$$
then we have
\begin{equation*}
p_{c}(f)\geq 1-\frac{\sqrt{d}}{\delta}(1-\rho_{\T}),
\end{equation*}
where $\ell$ is the cross entropy loss that $\ell(f(x),k)=-\ln(f_{k}(x))$, $\delta=\min(\delta_{0},\delta_{f}(e^{-L_{f}^{max}}-c))$, and $\delta_0$ is defined in Proposition~\ref{pro:separation_gap}.
\end{thm}
\begin{proof}
The proof can be found in ~\ref{proof:lower_bound}.
\end{proof}

Theorem~\ref{thm:lower_bound} reveals that the expected accuracy is related to the total cover $\rho_{\T}$, separation gap $\delta_0$, neural network smoothness $\delta_f$, and loss value $L_{f}^{max}$. We will show numerically in Section \ref{num_res} that $\delta_{f}(e^{-L_{f}^{max}}-c)$ increases first and then decreases during the training of neural networks. The following theorem states that the maximum value of $\delta_{f}(e^{-L_{f}^{max}}-c)$ is bounded by the empirical separation gap.

\begin{thm}[Empirical separation gap] \label{thm:deltaf_bound}
Let $f$ be a normalized continuous positive mapping. Suppose that $\T=\{x_{1},\cdots,x_{n}\}$ is a single label training set. For any $0.5\leq c<1$, if $L_{f}^{max}<-\ln c$, then we have
$$\delta_f(e^{-L_f^{max}}-c) \leq \delta_{\T}/2,$$
where
$$\delta_{\T} := \min_{tag(x_i) \neq tag(x_j)} \|x_i-x_j \| \geq \delta_0$$
is called the empirical separation gap, i.e., the smallest distance between two differently labeled training points.
\end{thm}
\begin{proof}
The proof can be found in ~\ref{proof:deltaf_bound}.
\end{proof}

Besides the upper bound, the lower bound of $\delta_f$ is also important to the accuracy. We have observed that in practice the low bound of $\delta_f$ exists (Figs. \ref{fig:one_two_dim}, \ref{fig:high_dim}, \ref{fig:depth_width} and \ref{fig:tr_data}), which indicates the existence of $\kappa$ in the following theorem. Based on this observation, we have the following theorem for the accuracy.

\begin{thm}[Lower bound of accuracy]  \label{thm:bounds}
Assume that there exists a constant $\kappa > 0$, such that
\begin{equation} \label{eq:assump}
    \kappa\delta_{0} \leq \kappa\delta_{\T} \leq \delta_{f}(e^{-L_f^{max}}-0.5) \leq \delta_{\T}/2
\end{equation}
holds for any single label training set $\T$ and a corresponding suitable trained network $f$ on $\T$ such that $L_{f}^{max}(\T)<\ln 2$, then we have the following conclusions for the expected accuracy $p(f)$ and the expected error $\E(f)=1-p(f)$:
\begin{enumerate}[(i)]
    \item with the same condition of Theorem~\ref{thm:lower_bound}, \\
    $p(f)\geq1-\frac{\sqrt{d}}{\delta}(1-\rho_{\T})$,
    \item $\E(f) \leq \frac{\sqrt{d}\cdot (1-\rho_{\T})}{\min(\delta_{0}, \kappa\delta_{\T})} = \alpha(\T)\cdot CC(\T),$
    \item $\lim\limits_{\rho_{\T}\to 1}p(f)=1,$
\end{enumerate}
where $\delta = \min(\delta_0, \delta_f(e^{-L_f^{max}}-c))$, and $\alpha(\T)=\frac{\sqrt{d}\cdot CD(\T)}{\min(\delta_{0}, \kappa\delta_{\T})}$.
\end{thm}
\begin{proof}
(\rnum 1) is the conclusion of Theorem \ref{thm:lower_bound}. The proof of (\rnum 2) and (\rnum 3) can be found in ~\ref{proof:bounds}.
\end{proof}

\begin{rem}
We have the following remarks for this main theorem:
\begin{itemize}
    \item In (\rnum 2), we rewrite $\frac{\sqrt{d}\cdot (1-\rho_{\T})}{\min(\delta_{0}, \kappa\delta_{\T})}$ to emphasize $CC(\T)$ by introducing a coefficient term $\alpha(\T)$. Although this seems artificial, our numerical experiments empirically show that in practice the linear scaling between $\E(f)$ and $CC(\T)$ is indeed satisfied.
    \item This theorem is not specific to neural networks, but rather holds for any trained model $f$ satisfying the assumptions required in the theorem. While the assumptions are not true for a general machine learning algorithm, we show numerically that in practice neural networks satisfy the assumptions.
    \item The current proof of the theorem relies on the assumption of the maximum cross entropy $L_f^{max}$, which could be hard to satisfy in practice, since it will be large even if only a single training sample is misclassified. However, the assumption of the maximum cross entropy is possible to be relaxed according to our experiments.
\end{itemize}
\end{rem}
Here, the cover complexity $CC(\T)$ consists of two parts, one represents the richness of the whole training set while the other part describes the degree of separation between different labeled subsets. As for $\alpha(\T)$, both the denominator and numerator seem to have a positive correlation with respect to separation level. What we wish is that $\alpha(\T)$ is almost close to a constant with high probability and the expected error $\E(f)$ is mainly determined by $CC(\T)$, which approximately represents the complexity level of the data set. We will provide more information in detail in the section concerning the numerical results.

\section{Numerical results}  \label{num_res}

In this section, we use numerical simulations to test the accuracy of neural networks in terms of the data distribution (cover complexity), and neural network smoothness. In addition, we study the effects of the network size and training dataset size on the smoothness. The codes are published in GitHub (https://github.com/jpzxshi/generalization).

\subsection{Data distribution}\label{sec:data}

\begin{table*}[htbp]
    \centering
    \begin{tabular}{cc|cc|ccc|cc}
        \toprule
        Data Set & Variants & Input dim ($d$) & Output dim ($K$) & $\rho_{\T}$ & $CD(\T)$ & $CC(\T)$ & $\E(f)$ & $\frac{\E(f)}{\sqrt{K}}$ \\
        \midrule
        MNIST & Original & 784 & 10 & .8480 & .1053 & 1.442 & .01 & .0032 \\
        CIFAR-10 & Original & 3072 & 10 & .8332 & .0163 & 10.23 & .45 & .1423 \\
        CIFAR-10 & Grey & 1024 & 10 & .8486 & .0125 & 12.11 & .53 & .1676 \\
        CIFAR-10 & Conv & 1024 & 10 & .9505 & .0094 & 5.280 & .18 & .0569 \\
        SVHN & Original & 3072 & 10 & .9034 & .0076 & 12.68 & .49 & .1550 \\
        SVHN & Grey & 1024 & 10 & .9117 & .0084 & 10.48 & .56 & .1771 \\
        SVHN & Conv & 1024 & 10 & .9632 & .0123 & 2.995 & .23 & .0727 \\
        \midrule
        CIFAR-100 & Original (coarse) & 3072 & 20 & .8337 & .0185 & 9.012 & .62 & .1386 \\
        CIFAR-100 & Grey (coarse) & 1024 & 20 & .8541 & .0132 & 11.08 & .72 & .1610 \\
        CIFAR-100 & Conv (coarse) & 1024 & 20 & .9626 & .0070 & 5.326 & .40 & .0894 \\
        COIL-20 & Original & 16384 & 20 & .9176 & .2385 & .3453 & .03 & .0067 \\
        \midrule
        CIFAR-100 & Original (fine) & 3072 & 100 & .8337 & .0270 & 6.149 & .73 & .0730 \\
        CIFAR-100 & Grey (fine) & 1024 & 100 & .8541 & .0198 & 7.380 & .81 & .0810 \\
        CIFAR-100 & Conv (fine) & 1024 & 100 & .9457 & .0136 & 4.000 & .52 & .0520 \\
        COIL-100 & Original & 49152 & 100 & .9430 & .1944 & .2930 & .01 & .0010 \\
        \bottomrule
    \end{tabular}
    \caption{Cover complexity $CC(\T)$, best error $\E(f)$, and normalized error $\frac{\E(f)}{\sqrt{K}}$ of different data sets. Different variants of data sets are used, including the original RGB or grey images (Original), grey images (Grey), and images extracted from a CNN (Conv). Images in CIFAR-100 have two variants: 100 categories (fine) and 20 categories (coarse). See~\ref{param_net} for details.}
    \label{tab:error_beta_table}
\end{table*}

In this subsection, we explore how $CC(\T)$ affects the expected error $\E(f)$. In our experiments, we test several data sets, including MNIST~\citep{lecun1998gradient}, CIFAR-10~\citep{krizhevsky2009learning}, CIFAR-100~\citep{krizhevsky2009learning}, COIL-20~\citep{nene1996columbia}, COIL-100~\citep{nene1996object}, SVHN~\citep{netzer2011reading}. In addition to the original data set, we also create some variants: (1) the images of grey color, (2) the images extracted from a convolutional layer after training the original data set using a convolutional neural network (CNN), (3) combine several categories into one category to reduce the number of total categories, see Table~\ref{tab:error_beta_table} and details in~\ref{param_net}.

For a training data set $\T$, we estimate $h_{\T}^{\mu}(r)$ by the proportion of the test data points within the balls with radius $r$ centered at training data points, i.e.,
$$h_{\T}^{\mu}(r) \approx \frac{\text{\# Test points within radius-$r$ balls of training points}}{\text{\# Test data points}},$$
and then $\rho_{\T}$ is obtained by Definition \ref{def:covers}. Similarly, we estimate $CD(\T)$ and then compute $CC(\T)$. Next for each data set, we train fully-connected neural networks with different hyperparameters, and record the best error we observed, see the details in~\ref{param_net}. The cover complexity and the best error for each data set are shown in Table~\ref{tab:error_beta_table}.

\begin{figure}[htbp]
    \centering
    \includegraphics[width=0.47\textwidth]{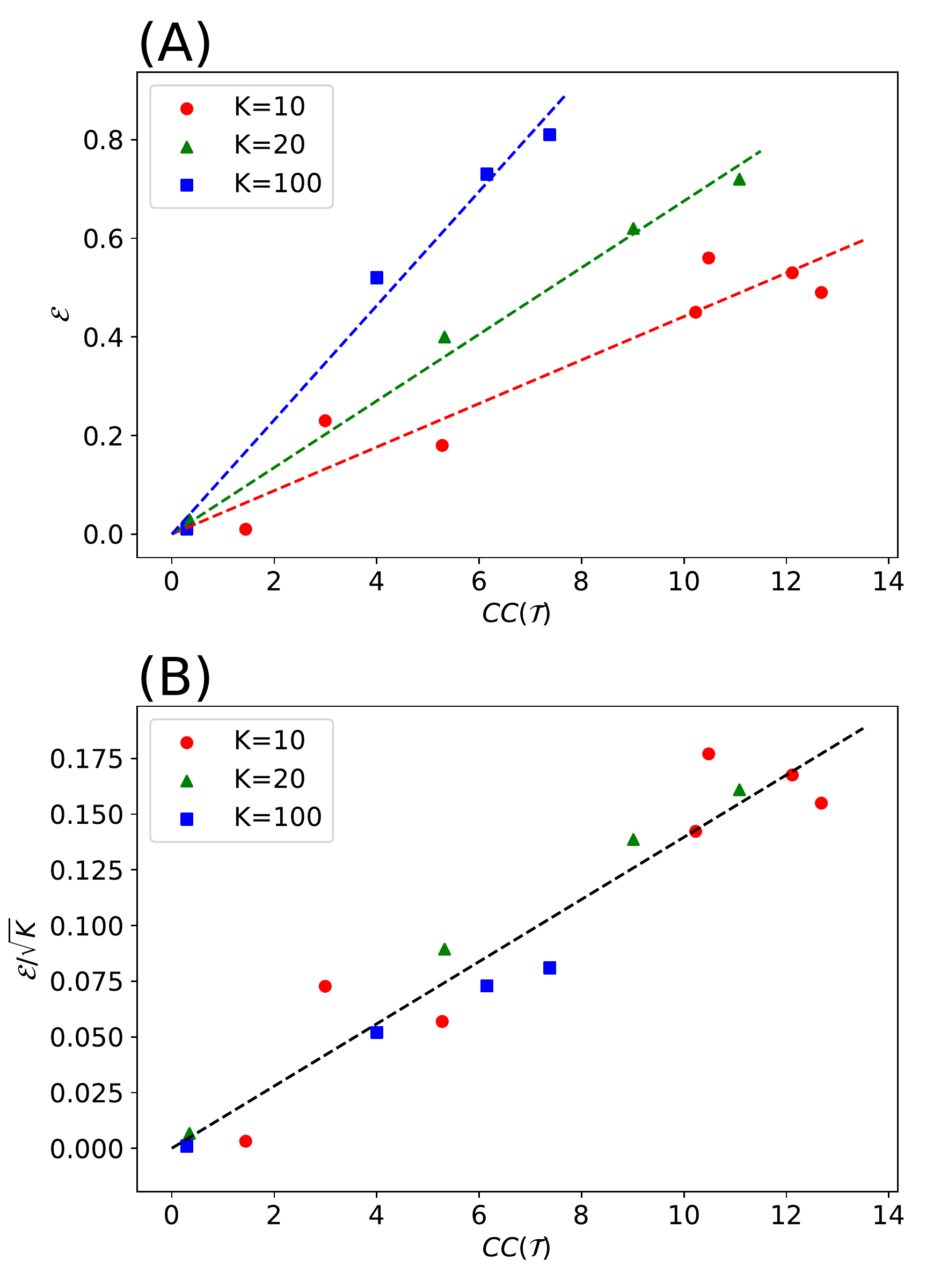}
    \caption{Relationship between cover complexity and error achieved by fully-connected neural networks. (\textbf{A}) Linear relationship between cover complexity $CC(\T)$ and error $\E(f)$ for different data sets with different category number $K$. The coefficients of determination of the linear regression (with the constraint that the line must pass through the origin) are $R^{2} \approx 0.89,0.99,0.98$ for $K=10,20,100$, respectively. (\textbf{B}) Linear relationship between cover complexity $CC(\T)$ and normalized error $\frac{\E(f)}{\sqrt{K}}$ for different data sets. Red, green and blue points represent data sets with output dimension of 10, 20 and 100, respectively. The line is fitted as $\frac{\E(f)}{\sqrt{K}} \approx 0.014 CC(\T)$, and the coefficient of determination is $R^{2} \approx 0.92$.}
    \label{fig:error_CC}
\end{figure}

These data sets are divided into three groups according to their output dimensions. For each group of the same output dimension, the error is almost linearly correlated with $CC(\T)$, see Fig.~\ref{fig:error_CC}A, regardless of the input dimension. In addition, we find that all the cases collapse into a single line when normalizing the error $\E(f)$ by a factor of $1/\sqrt{K}$, see Fig.~\ref{fig:error_CC}B.

It is noteworthy that the $CC(\T)$ of convolutional variants of data sets is much smaller than that of the original data sets, and hence the expected accuracy increases. The results confirm the importance of data distribution.

\begin{figure}[htbp]
    \centering
    \includegraphics[width=0.5\textwidth]{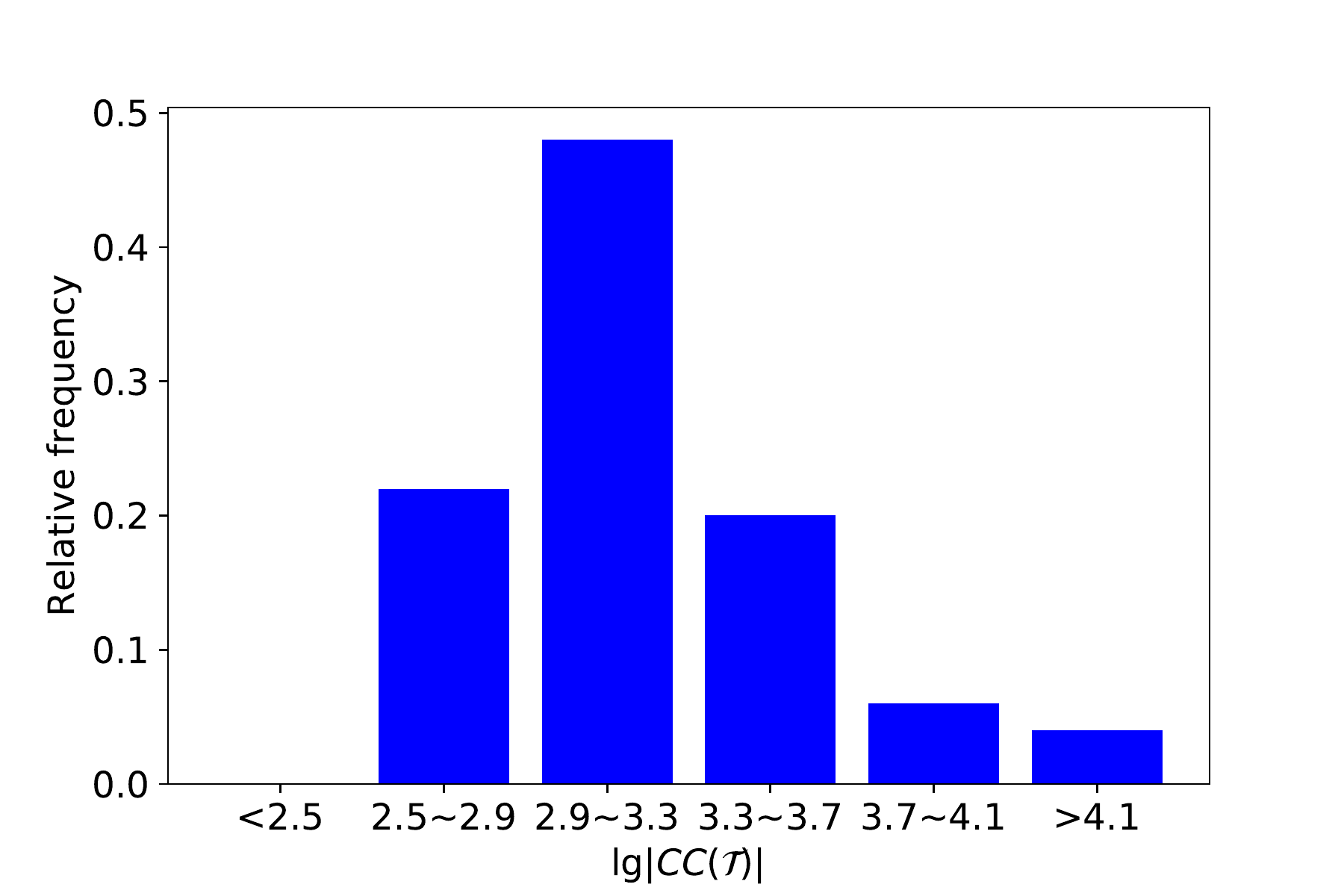}
    \caption{Distribution of $CC(\T)$ of randomly labeled MNIST data set. We assign each image in MNIST a random label and compute its $CC(\T)$. The smallest $|CC(\T)|$ is $\sim$300. The distribution is obtained from 50 random data sets.}
    \label{fig:random_CC}
\end{figure}

Next, we consider the most difficult data set, i.e., data with random labels. We choose MNIST and then assign each image a random label. We repeat this process 50 times, and compute each $CC(\T)$. The distribution of $|CC(\T)|$ is shown in Fig.~\ref{fig:random_CC}. The smallest $|CC(\T)|$ is $\sim$300, which is much larger than that of the original data sets with $CC(\T) < 20$. This extreme example again confirms that $CC(\T)$ is a proper measure of the difficulty of classifying a data set.

\subsection{Neural network smoothness}

In this subsection, we will investigate the relationship between the neural network smoothness $\delta_f(e^{-L_f^{max}}-c)$ and the accuracy, and the effects of network size (depth and width) on the smoothness. We first show results for one- and two-dimensional problems, where $\delta_f(e^{-L_f^{max}}-c)$ can be computed accurately by brute force. Subsequently, we consider the high dimensional setting of the MNIST data set, where we estimate $\delta_f(e^{-L_f^{max}}-c)$ by Eq.~(\ref{eq:lip}).

\subsubsection{One- and two-dimensional problems}

We first consider a one-dimensional case and a two-dimensional case. For the one-dimensional case, we choose the sample space $D = [0, 1]$, $K=2$, and the ideal label function as
$$tag(x) = \begin{cases} \{1\} & \text{if } x \in [0, 0.5-\frac{\delta_0}{2}] \\ \{1,2\} & \text{if } x \in (0.5-\frac{\delta_0}{2}, 0.5+\frac{\delta_0}{2}) \\ \{2\} & \text{if } x \in [0.5+\frac{\delta_0}{2}, 1] \end{cases},$$
with separation gap $\delta_{0}=0.1$. We use $n$ equispaced points ($n \geq 4$ is an even number) on $D\setminus (0.5-\frac{\delta_0}{2}, 0.5+\frac{\delta_0}{2})$ as the training set, i.e., $\T = \T_{1} \cup \T_{2}$, where \begin{equation*}
\begin{split}
&\T_{1}= \left\{0, \frac{1-\delta_0}{n-2}, \frac{1-\delta_0}{n-2}\cdot 2, \dots, \frac{1}{2}-\frac{\delta_0}{2}\right\}, \\
&\T_{2}= \left\{\frac{1}{2}+\frac{\delta_0}{2}, \dots, 1-\frac{1-\delta_0}{n-2}, 1\right\}.
\end{split}
\end{equation*}
We choose 10000 equispaced points on $D$ as the test data.

For the two-dimensional case, we choose the sample space $D = [0, 1]^2$, $K=2$, and the ideal label function as
\begin{equation*}
tag(\mathbf{x}) = \begin{cases}
\{1\} & \text{if } \|\mathbf{x} - (0.5, 0.5)\| \leq 0.4 - \frac{\delta_0}{2} \\
\{2\} & \text{if } \|\mathbf{x} - (0.5, 0.5)\| \geq 0.4 + \frac{\delta_0}{2} \\
\{1,2\} & \text{otherwise}
\end{cases},
\end{equation*}
with $\delta_{0}=0.1$. For the training set, we first choose $n = m^2$ equispaced points, i.e., $\T= \{0, \frac{1}{m-1}, \frac{2}{m-1}, \dots, \frac{m-2}{m-1}, 1\}^2$, and then remove the points with label $\{1,2\}$ to ensure that all samples are of single label. We choose $1000^{2}$ equispaced points on $D$ as the test data.

In our experiments, we use a 3-layer fully-connected NN with ReLU activation and 30 neurons per layer. The neural network is trained for 1000 iterations by the Adam optimizer~\citep{kingma2014adam} for the one-dimensional problem, and 2000 iterations for the two-dimensional problem. For the one-dimensional problem, the $c$-accuracy $p_c(f)$ with $c=0.5$ and lower bounds for different numbers of training points are listed in Table~\ref{tab:case_onedim}. We can see that the bounds become tighter when $n$ is larger.

\begin{table}[htbp]
    \centering
    \begin{tabular}{c|c|cc|ccc}
        \toprule
        $n$ & $L_f^{max}$ & $\rho_{\T}$ & $\delta$ & $p_c(f)$ & $h_{\T}(\delta)$ & $1 - \frac{\sqrt{d}}{\delta}(1-\rho_{\T})$ \\
        \midrule
        10 & .285 & .972 & .045 & 1.0 & 0.80 & 0.38 \\
        20 & .246 & .988 & .041 & 1.0 & 1.00 & 0.69 \\
        40 & .182 & .994 & .041 & 1.0 & 1.00 & 0.85 \\
        80 & .127 & .997 & .038 & 1.0 & 1.00 & 0.92 \\
        \bottomrule
    \end{tabular}
    \caption{Comparison between $c$-accuracy $p_c(f)$ with $c=0.5$ and the lower bounds for different training set sizes for the one-dimensional problem. Here $\delta$ is indicated in Theorem~\ref{thm:lower_bound}. The neural network is trained for 1000 iterations by the Adam optimizer.}
    \label{tab:case_onedim}
\end{table}

\begin{figure}[htbp]
    \centering
    \includegraphics[width=0.47\textwidth]{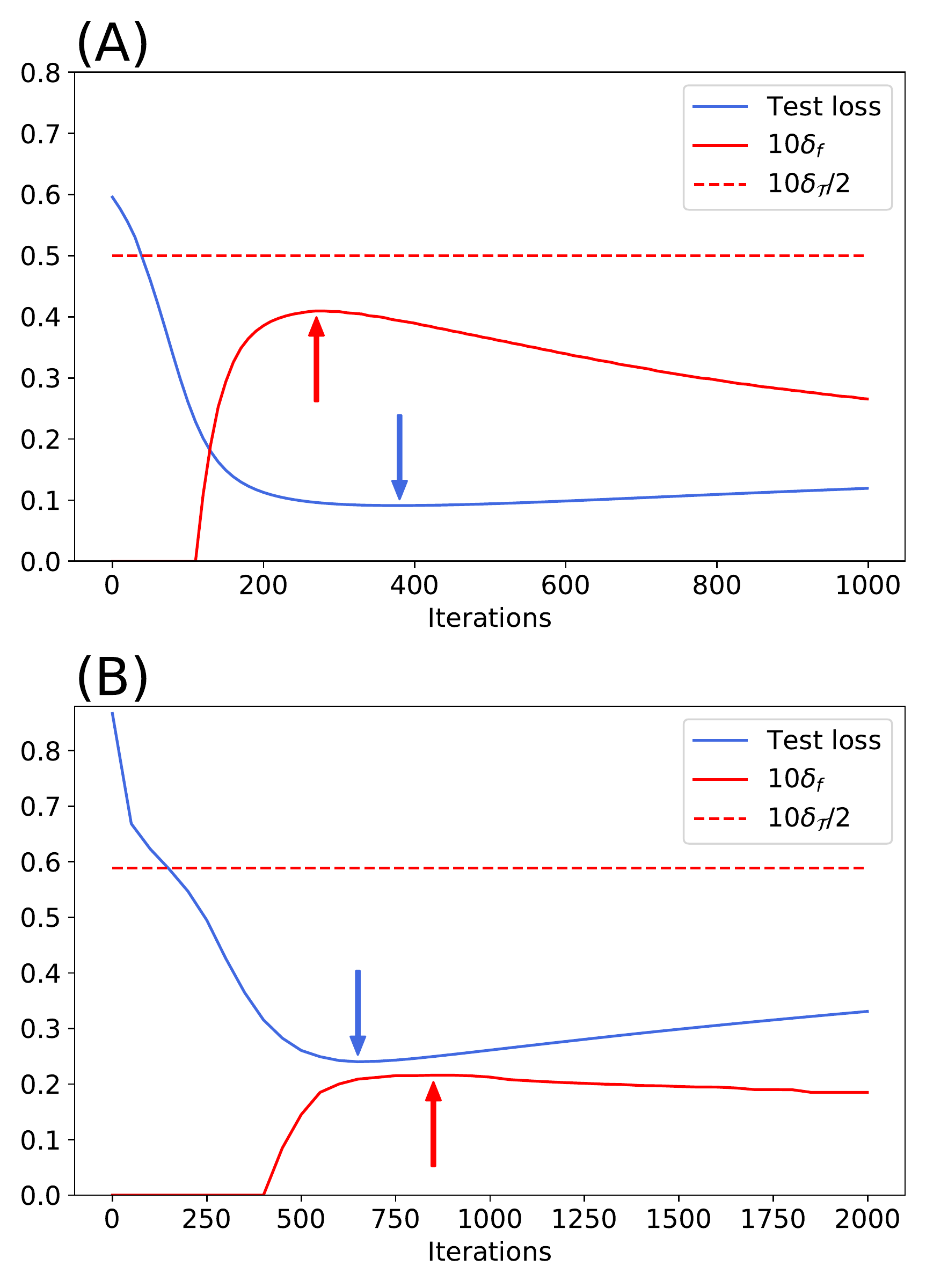}
    \caption{Consistency between the test loss and neural network smoothness $\delta_{f}(e^{-L_{f}^{max}}-\frac{1}{2})$ during the training process of the neural network. (\textbf{A}) One-dimensional problem of $n=20$. (\textbf{B}) Two-dimensional problem of $n=400$. $\delta_f$ is bounded by $\delta_{\T}/2$. The arrows indicate the minimum of the test loss and the maximum of $\delta_f$.}
    \label{fig:one_two_dim}
\end{figure}

During the training process of the neural network, the test loss first decreases and then increases, while $\delta_{f}$ first increases and then decreases, see Fig.~\ref{fig:one_two_dim}A for the one-dimensional problem ($n=20$) and Fig.~\ref{fig:one_two_dim}B for the two-dimensional problem ($n=400$). $\delta_f$ is bounded by $\delta_{\T}/2$, as proved in Theorem~\ref{thm:deltaf_bound}. We also observe that the trends of test loss and $\delta_f$ coincide, and thus we should stop the training when $\delta_f$ begins to decrease to prevent overfitting.

\subsubsection{High-dimensional problem}

\begin{figure}[htbp]
    \centering
    \includegraphics[width=0.5\textwidth]{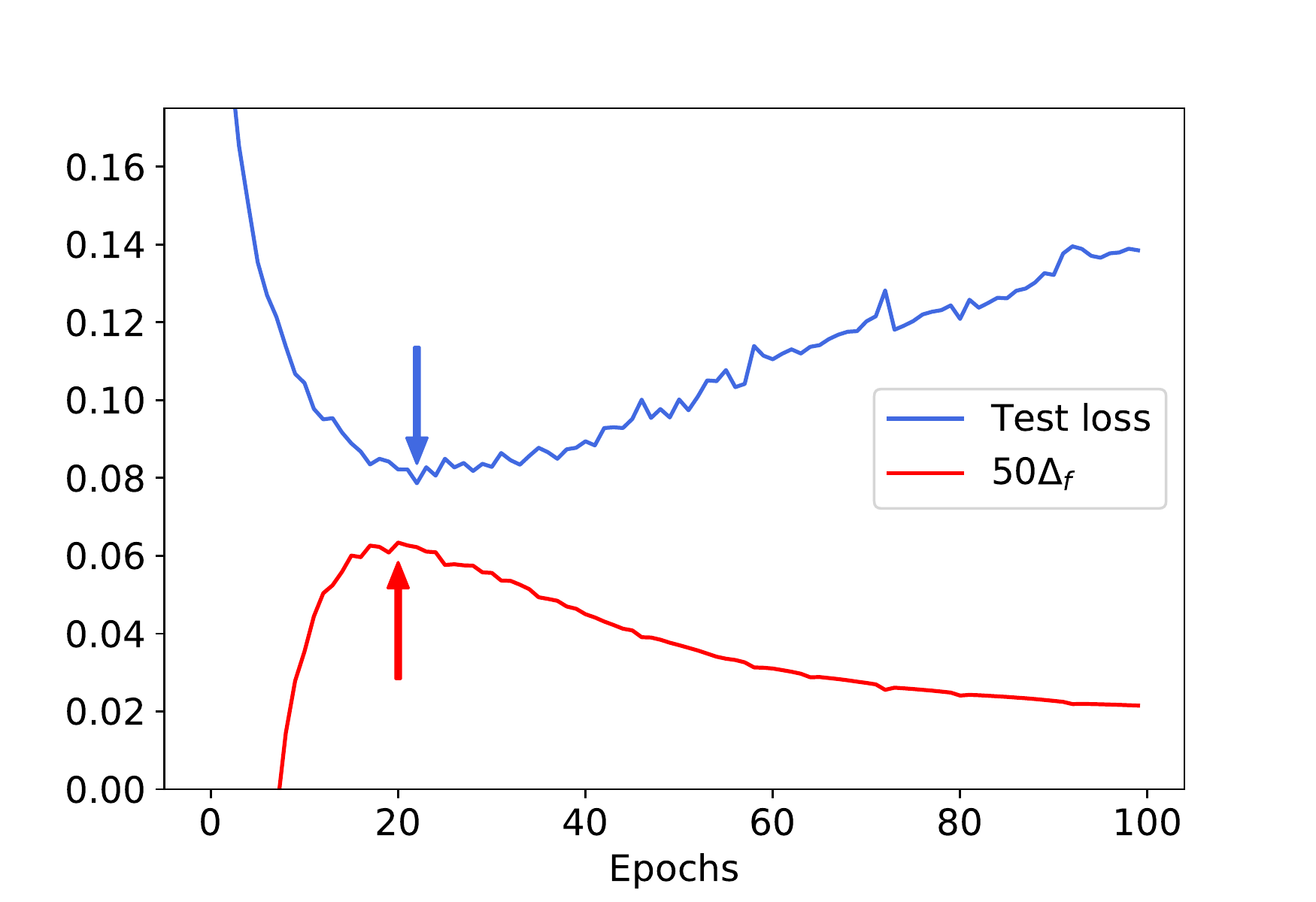}
    \caption{Consistency between test loss and neural network smoothness during the training of the neural network for MNIST. The arrows indicate the minimum of the test loss and the maximum of $\Delta_f$.}
    \label{fig:high_dim}
\end{figure}

In the high-dimensional problem of MNIST, we consider the average loss $L_f$ instead of the maximum loss $L_f^{max}$, which is very sensitive to extreme points. As shown in Eq.~(\ref{eq:lip}), we use the following quantity to bound $\delta_f(e^{-L_f}-c)$:
$$\Delta_{f}=\frac{e^{-L_{f}}-c}{\|W_{1}\|_{2}\cdots\|W_{l}\|_{2}}.$$
Because we use the $c$-accuracy to approximate the true accuracy, for the classification problems with two categories, $p_{0.5}(f)$ and $p(f)$ are equivalent. However, they are not equal for problems with more than two categories, where the best $c$ depends on the properties of the data set, such as the easiness of learning to classify the data set. If the data set is easy to classify, such as MNIST, the best $c$ should be close to 1. In our example, we choose $c=0.9$. We train MNIST using a 3-layer fully-connected NN with ReLU activation and 100 neurons per layer for 100 epochs. In Fig.~\ref{fig:high_dim}, we can also see the consistency between the test loss and neural network smoothness, as we observed in the low-dimensional problems.

\subsubsection{Effects of the network size and training dataset size on the smoothness}

\begin{figure}[htbp]
    \centering
    \includegraphics[width=0.48\textwidth]{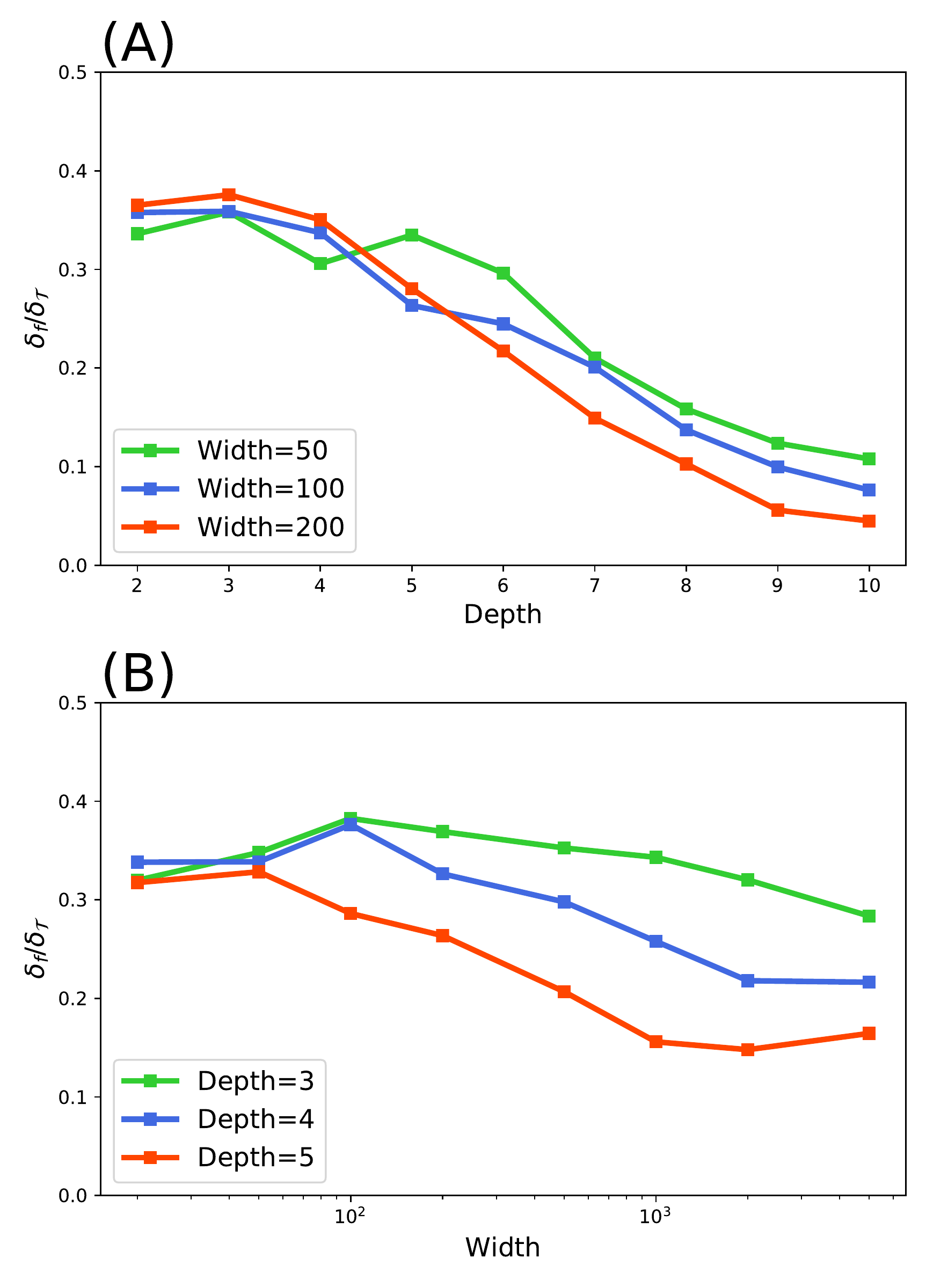}
    \caption{Effects of network depth and width on the normalized smoothness $\delta_{f}/\delta_{\T}$. Recall that $\delta_{f}=\delta_{f}(e^{-L_{f}^{max}}-\frac{1}{2})$. (\textbf{A}) The $\delta_{f}/\delta_{\T}$ decreases fast when the network depth increases. (\textbf{B}) The $\delta_{f}/\delta_{\T}$ decreases relatively slow when the network width increases. The results are averaged from 50 independent experiments.}
    \label{fig:depth_width}
\end{figure}

We have demonstrated that network smoothness $\delta_{f}(e^{-L_{f}^{max}}-c)$ is an important factor to the accuracy. Next, we investigate the effects of network size (depth and width) on the smoothness for binary classification problems, which are explained as follows. We consider the one-dimensional sample space $D=[0, 1]$, and choose $n$ equispaced points on $D$ as the training data locations. To avoid the effects of the choice of target true functions, we always repeat experiments with different target functions, and in each experiment we generate a random target function. Specifically, to generate a random target function, we first sample two random functions $g_1(x)$ and $g_2(x)$ from a Gaussian process with the radial basis function kernel of a length scale 0.2, and then assign a point $x$ as category 1 if $g_1(x) > g_2(x)$, otherwise assign this point as category 2. When  training neural networks, we monitor the value of $\delta_{f}(e^{-L_{f}^{max}}-c)$ and stop the training once $\delta_{f}$ begins to decrease as shown in Figs.~\ref{fig:one_two_dim} and \ref{fig:high_dim}. We first choose the dataset size $n=10$, and we show that the normalized smoothness $\delta_{f}/\delta_{T}$ decreases as the network depth or width increases (Fig.~\ref{fig:depth_width}). We also show that the effects of depth is more significant than that of width.

\begin{figure}[htbp]
    \centering
    \includegraphics[width=0.48\textwidth]{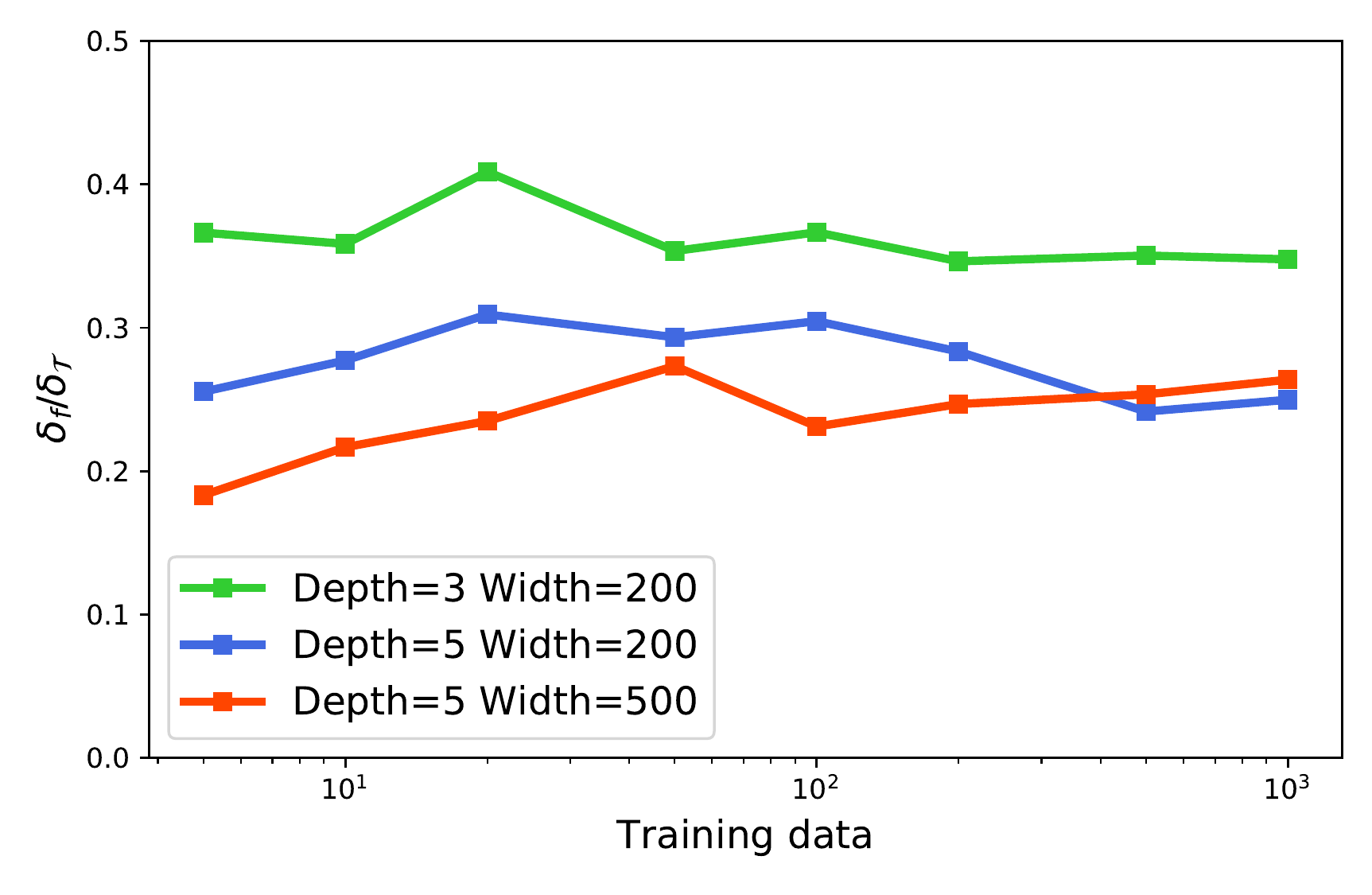}
    \caption{Effects of the training data size on the normalized $\delta_{f}/\delta_{\T}$. Recall that $\delta_{f}=\delta_{f}(e^{-L_{f}^{max}}-\frac{1}{2})$. The $\delta_{f}/\delta_{\T}$ is insensitive to the training data size, and is bounded from below by a positive constant. The results are averaged from 50 independent experiments.}
    \label{fig:tr_data}
\end{figure}

Our main theorem (Theorem \ref{thm:bounds}) requires the assumption (Eq. \ref{eq:assump}), which would not be true for a general machine learning algorithms. Here, we verify this assumption for neural networks by numerical experiments. Specifically, we train a fully-connected neural networks using training datasets of different size $n$. We show that $\delta_{f}/\delta_{\T}$ is insensitive to training dataset size, and is always bounded from below by a positive constant $\kappa$ (Fig.~\ref{fig:tr_data}). This result reveals that the neural networks would fit a dataset in a relatively smooth way during the training process.

\section{Discussion} \label{disc}

When neural networks are used to solve classification problems, we expect that the accuracy is dependent on some properties of the data set. It is still quite surprising, however, that there is a linear relationship between the accuracy and the cover complexity of the data set, as we have seen in Section~\ref{sec:data}. Theorem~\ref{thm:bounds}(ii) provides an upper bound of the error, but a lower bound is missing. To fully explain this observation, two conjectures of the learnability of fully-connected neural networks are proposed: when a neural network $f$ is trained on a data set $\T$ in such a way that $L_{f}^{max}(\T)<\ln 2$ and $\delta_f(e^{-L}-c) \asymp \delta_{\T}$, then we have
\begin{itemize}
\item $\E(f) \approx c(K) \cdot CC(\T)$,
  where $c(K)$ is a constant depending only on $K$.
\item $\frac{\E(f)}{\sqrt{K}} \approx c \cdot CC(\T)$,
  where $c \approx 0.014$ is a constant.
\end{itemize}

On the other hand, the theoretical and numerical results provide a better understanding of the generalization of neural network from the training procedure. The smoothness $\delta_f(e^{-L}-c)$ of neural networks plays a key role, where $L$ is the maximum cross entropy loss $L_f^{max}$ or the average cross entropy loss $L_f$. We can see that:
\begin{itemize}
    \item $\delta_f(e^{-L}-c)$ depends on both the regularity of $f$ and the loss value $L$ (which also depends on $f$). Large $\delta_f(e^{-L}-c)$ requires good regularity and large $e^{-L}-c$, i.e., small $L$. However, small $L$ could correspond to bad regularity of $f$. Thus, there is a trade-off between the loss value $L$ and the regularity of $f$.
    \item Due to this trade-off, $\delta_f(e^{-L}-c)$ increases first and then decrease during the training process. Hence, we should not optimize neural networks excessively. Instead, we should stop the training early when $\delta_f(e^{-L}-c)$ begins to decrease, which leads to another ``early stopping'' strategy to prevent overfitting. 
\end{itemize}

We also note that the lower bound of $\delta_f(e^{-L}-c)$ in Eq.~(\ref{eq:lip}) relates to the norm of weight matrices of neural networks:
$$\delta_{f}(e^{-L}-c)\gtrsim\frac{e^{-L}-c}{\|W_{1}\|_{2}\cdots\|W_{l}\|_{2}}.$$
There have been some works to study the norm-based complexity of neural networks (see the Introduction), and these bounds typically scale with the product of the norms of the weight matrices, e.g.,~\citep{neyshabur2017exploring}
\begin{equation*}
\frac{1}{\gamma_{margin}^{2}}\prod_{i=1}^{l}h_{i}\|W_{i}\|_{2}^{2},
\end{equation*}
where $h_{i}$ and $W_{i}$ are the number of nodes and the weight matrix in layer $i$ of a network with $l$-layers, and $\gamma_{margin}$ is the margin quantity, which describes the goodness of fit of the trained network to the data. The product of the matrix norms depends exponentially on the depth, while some recent works show that the generalization bound could scale polynomially in depth under some assumptions~\citep{nagarajan2018deterministic,wei2019data}. The exploration of the dependence of $\delta_f(e^{-L}-c)$ on depth is left for future work.

\section{Conclusion} \label{conc}

In this paper, we study the generalization error of neural networks for classification problems in terms of the data distribution and neural network smoothness. We first establish a new framework for classification problems. We introduce the \textit{cover complexity} (CC) to measure the difficulty of learning a data set, an accuracy measure called $c$-\textit{accuracy} which is stronger than the standard classification accuracy, and \textit{the inverse of the modulus of continuity} to quantify neural network smoothness. Subsequently, we derive a quantitative bound for the expected accuracy/error in Theorem~\ref{thm:bounds}, which considers both the cover complexity and neural network smoothness.

We validate our theoretical results by several data sets of images. Our numerical results demonstrate that CC is a reliable measure for the difficulty of learning to classify a data set. On the other hand, we observe a clear consistency between test loss and neural network smoothness during the training process. We also show that neural network smoothness decreases when the network depth and width increases, and the effects of depth is more significant than that of width, while the smoothness is insensitive to training dataset size.

\section*{Acknowledgements}

This work is supported by the DOE PhILMs project (No. de-sc0019453), the AFOSR grant FA9550-17-1-0013, and the DARPA AIRA grant HR00111990025. The work of P. Jin and Y. Tang is partially supported by the Major Project on New Generation of Artificial Intelligence from MOST of China (Grant No. 2018AAA0101002), and the National Natural Science Foundation of China (Grant No. 11771438).

\appendix

\section{Example of topology} \label{exam:topo}
\begin{exa}
Given $$T=2^{\{1,2,3\}}\backslash \{\varnothing\}=\{\ \{1\},\ \{2\},\ \{3\},\ \{1,2\},\ \{1,3\},\ \{2,3\},\ \{1,2,3\}\ \},$$
\begin{equation*}
\begin{split}
U=\{\
 &\{ \{1\},\{1,2\},\{1,3\},\{1,2,3\}  \},  \   \{  \{2\},\{2,1\},\{2,3\},\{1,2,3\}  \},   \\
 &\{  \{3\},\{3,1\},\{3,2\},\{1,2,3\}  \},  \  \{  \{1,2\},\{1,2,3\}  \},  \\    
 &\{  \{1,3\},\{1,2,3\}  \}, \  \{  \{2,3\},\{1,2,3\}  \}, \ \{ \{1,2,3\}\}\ \},
\end{split}
\end{equation*}
then $\tau_{T}$ is the topology generated by $U$.
\end{exa}

In this example, $\{ \{1\},\{1,2\},\{1,3\},\{1,2,3\}  \}$ is an open set, since it consists of all elements containing label $\{1\}$, and $\{  \{2,3\},\{1,2,3\}  \}$ is also an open set with common part $\{2,3\}$. Besides open sets from base $U$, $\{\{1\},\{1,2\},\{1,3\},\{2,3\},\{1,2,3\}\}$ is still an open set as the union of the two shown above.

\section{Proof of Proposition \ref{pro:separation_gap}} \label{proof1}
\begin{proof}
We use the proof by contradiction. Assume that the result does not hold, then
\begin{center}
$\exists\{x_{n}\},\{y_{n}\}\subseteq D,\ s.t.\ \|x_{n}-y_{n}\|\rightarrow 0(n\rightarrow\infty)$,\ and $tag(x_{n})\cap tag(y_{n})=\varnothing.$\\
\end{center}
Because $D$ is compact, there exist $x^{*}\in D$ and a subsequence $\{x_{k_{n}}\}$ of $\{x_{n}\}$ such that $\ x_{k_{n}}\rightarrow x^{*}.$ As $\|x_{n}-y_{n}\|\rightarrow 0$, also $\ y_{k_{n}}\rightarrow x^{*}.$ Choose any $i_{0}\in tag(x^{*})$, then there exists a sufficient large $k_{n_0}$ such that $\ i_{0}\in tag(x_{k_{n_{0}}}),\ i_{0}\in tag(y_{k_{n_{0}}}).$ Therefore $tag(x_{k_{n_{0}}})\cap tag(y_{k_{n_{0}}})\neq \varnothing$, which contradicts the assumption.
\end{proof}

\section{Proof of Proposition \ref{pro:geo_interpretation}} \label{proof:geo_interpretation}
\begin{proof}
Let $\delta_{0}$ as defined in this proposition. For any two different points $x,y$ with distance less than $\delta_{0}$, either $tag(x)=tag(y)$, or at least one of the two is a full label point, in both cases $tag(x)\cap tag(y)\neq \varnothing$. For any $\delta>\delta_{0}$, according to the definition of $\delta_{0}$, there exist two points $x_{0},y_{0}$ satisfying
\begin{equation*}
\|x_{0}-y_{0}\|<\delta,\quad tag(x_{0})\cap tag(y_{0})=\varnothing.
\end{equation*}
The two facts imply that $\delta_{0}$ is the supremum of $\delta$ satisfying Proposition \ref{pro:separation_gap}.
\end{proof}

\section{Proof of Proposition \ref{pro:h_estimation}} \label{proof3}
\begin{proof}
According to the definition,
\begin{equation*}
\begin{split}
\sqrt{d}\rho_{\T}&=\int_{0}^{\sqrt{d}}h_{\T}^{\mu}(t)dt \\
&=\int_{0}^{r}h_{\T}^{\mu}(t)dt+\int_{r}^{\sqrt{d}}h_{\T}^{\mu}(t)dt \\
&\leq r\cdot h_{\T}^{\mu}(r)+(\sqrt{d}-r)\cdot 1 \\
&=\sqrt{d}-r(1-h_{\T}^{\mu}(r)),
\end{split}
\end{equation*}
thus
$$h_{\T}^{\mu}(r)\geq 1-\frac{\sqrt{d}}{r}(1-\rho_{\T}).$$
\end{proof}

\section{Estimate of total cover} \label{ETC}

In this section, we estimate the TC by the number of samples in the training set. The notations, such as $D$, $d$, $\mu$, $\rho_{\T}$, as well as training set
\begin{equation*}
\T=\{x_{1},x_{2},\dots,x_{n}\}\subseteq D
\end{equation*}
are the same as before. Note that samples in $\T$ are drawn according to $\mu$. Before presenting the analysis, we first collect the following auxiliary notions and results ({\bf Definitions \ref{def1}-\ref{def4}, Theorem \ref{epsilon_net}}) which are easily found in \citet{michael2017probability} (Definitions 14.1-14.3, Definition 14.5, and Theorem 14.8):

\begin{defi}\label{def1}
A range space is a pair $(X,\mathcal{R})$ where:
\begin{enumerate}
\item $X$ is a (finite or infinite) set of points;
\item $\mathcal{R}$ is a family of subsets of $X$, called ranges.
\end{enumerate}
\end{defi}

\begin{defi}
Let $(X,\mathcal{R})$ be a range space and let $Y \subseteq X$. The projection of $\mathcal{R}$ on
$Y$ is
$$\mathcal{R}_{Y} = \{R\cap Y | R\in \mathcal{R}\}.$$
\end{defi}

\begin{defi}
Let $(X,\mathcal{R})$ be a range space. A set $Y \subseteq X$ is shattered by $\mathcal{R}$ if
$|\mathcal{R}_{Y}| = 2^{|Y|}$. The Vapnik-Chervonenkis (VC) dimension of a range space $(X,\mathcal{R})$ is
the maximum cardinality of a set $Y \subseteq X$ that is shattered by $\mathcal{R}$. If there are arbitrarily
large finite sets that are shattered by $\mathcal{R}$, then the VC dimension is infinite.
\end{defi}

\begin{defi}\label{def4}
Let $(X,\mathcal{R})$ be a range space, and let $\mathcal{F}$ be a probability distribution
on $X$. A set $N \subseteq X$ is an $\epsilon$-net for $X$ with respect to $\mathcal{F}$ if for any set $R\in \mathcal{R}$ such that
$Pr_{\mathcal{F}}(R)\geq\epsilon$, the set $R$ contains at least one point from $N$, i.e.,
$$\forall R\in \mathcal{R}, Pr_{\mathcal{F}}(R)\geq \epsilon \Rightarrow R\cap N \neq \varnothing.$$
\end{defi}

\begin{thm}\label{epsilon_net}
Let $(X,\mathcal{R})$ be a range space with VC dimension $d_{vc}$ and let $\mathcal{F}$ be a probability distribution on $X$. For any $0<\eta,\epsilon\leq1/2$, there is an
$$m=O\left(\frac{d_{vc}}{\epsilon}\ln\frac{d_{vc}}{\epsilon}+\frac{1}{\epsilon}\ln\frac{1}{\eta}\right)$$
such that a random sample from $\mathcal{F}$ of size greater than or equal to $m$ is an $\epsilon$-net for $X$
with probability at least $1-\eta$.
\end{thm}

Now let
\begin{equation*}
B_{D}=\{B(x,r)\cap D | x\in \R^{d}, r>0\},
\end{equation*}
we first show $(D,B_{D})$ is a range space with VC dimension $d+1$.
\begin{lemma}
The VC dimension of range space $(D,B_{D})$ is $d+1$.
\end{lemma}
\begin{proof}
The proof can be found in \citet{dudley1979balls}.
\end{proof}
Set
\begin{equation*}
B_{D}^{\epsilon}(r)=\{A\in B_{D}|A=B(x,s)\cap D, 0<s\leq r, \mu(A)\geq\epsilon\},
\end{equation*}
and
\begin{equation*}
\varrho(\epsilon)=\frac{1}{\sqrt{d}}\int_{0}^{\sqrt{d}}\mu\left(\bigcup\limits_{A\in B_{D}^{\epsilon}(\frac{1}{2}r)}A\right)dr,
\end{equation*}
we have the following lemmas.
\begin{lemma}\label{lem7}
$\bigcup\limits_{A\in B_{D}^{\epsilon}(\frac{1}{2}r)}A\subseteq D\cap\bigcup\limits_{x_{i}\in \T}B(x_{i},r)$ when $\T$ is an $\epsilon$-net for $(D,B_{D})$.
\end{lemma}
\begin{proof}
For any $x\in\bigcup\limits_{A\in B_{D}^{\epsilon}(\frac{1}{2}r)}A$, we have $x\in A^{*}=B(x^{*},s^{*})\cap D$ for certain $s^{*}\leq \frac{1}{2}r$, with $\mu(A^{*})\geq\epsilon$. Since $\T$ is an $\epsilon$-net and $\mu(A^{*})\geq\epsilon$, we know $\T\cap A^{*}\neq \varnothing$. Thus there exists $x_{t}\in \T$ such that $\|x_{t}-x^{*}\|<s^{*}\leq\frac{1}{2}r$. Therefore
\begin{equation*}
\|x-x_{t}\|\leq\|x-x^{*}\|+\|x^{*}-x_{t}\|<\frac{1}{2}r+\frac{1}{2}r=r.
\end{equation*}
The above inequality shows that $$x\in  D\cap B(x_{t},r)\subseteq D\cap\bigcup\limits_{x_{i}\in \T}B(x_{i},r).$$
\end{proof}

\begin{lemma}\label{lem8}
$\lim\limits_{\epsilon\to 0}\varrho(\epsilon)=1$.
\end{lemma}
\begin{proof}
For any positive decreasing sequence $\{\epsilon_{i}\}$ which satisfies $\lim_{i\to \infty}\epsilon_{i}=0$, it leads to an increasing chain 
$$\left(\bigcup\limits_{A\in B_{D}^{\epsilon_{i}}(\frac{1}{2}r)}A\right)\subseteq\left(\bigcup\limits_{A\in B_{D}^{\epsilon_{i+1}}(\frac{1}{2}r)}A\right),\quad \lim_{i\to \infty}\left(\bigcup\limits_{A\in B_{D}^{\epsilon_{i}}(\frac{1}{2}r)}A\right)=\widetilde{D}\subseteq D,$$
where $\widetilde{D}=\bigcup_{A\in B_{D}^{+}(\frac{1}{2}r)}A$ for $B_{D}^{+}(\frac{1}{2}r)=\{A\in B_{D}|A=B(x,s)\cap D, 0<s\leq \frac{1}{2}r, \mu(A)> 0\}$. Let us consider a series of open balls $\{B_{i}\}$ of radius at most $\frac{1}{2}r$ that cover $D$, and we divide them into two parts $\{B_{i}\}=\{B_{i}^{+}\}\cup\{B_{i}^{0}\}$ such that $\mu(B_{i}^{+}\cap D)>0$ and $\mu(B_{i}^{0}\cap D)=0$. Then $\mu(\widetilde{D})\geq\mu(D\cap\bigcup_{i}B_{i}^{+})\geq\mu(D)-\mu(D\cap\bigcup_{i}B_{i}^{0})=1-0=1$, and thus $\mu(\widetilde{D})=1$. Therefore, we have $\lim_{\epsilon\rightarrow0}\mu(\bigcup_{A\in B_{D}^{\epsilon}(\frac{1}{2}r)}A)=\mu(\widetilde{D})=1$.

Since
\begin{equation*}
\lim_{\epsilon\rightarrow0}\mu\left(\bigcup\limits_{A\in B_{D}^{\epsilon}(\frac{1}{2}r)}A\right)=1\ and\ \mu\left(\bigcup\limits_{A\in B_{D}^{\epsilon}(\frac{1}{2}r)}A\right)\leq 1,
\end{equation*}
by dominated convergence theorem, we have
\begin{equation*}
\begin{split}
\lim_{\epsilon\rightarrow0}\varrho(\epsilon)&=\lim_{\epsilon\rightarrow0}\frac{1}{\sqrt{d}}\int_{0}^{\sqrt{d}}\mu\left(\bigcup\limits_{A\in B_{D}^{\epsilon}(\frac{1}{2}r)}A\right)dr \\
&=\frac{1}{\sqrt{d}}\int_{0}^{\sqrt{d}}\lim_{\epsilon\rightarrow0}\mu\left(\bigcup\limits_{A\in B_{D}^{\epsilon}(\frac{1}{2}r)}A\right)dr \\
&=\frac{1}{\sqrt{d}}\int_{0}^{\sqrt{d}}1dr \\
&=1.
\end{split}
\end{equation*}
\end{proof}
According to the aforementioned lemmas, we deduce the following theorem.
\begin{thm}\label{thm3}
Let $\T=\{x_{1},x_{2},\dots,x_{n}\}$ be the training set drawn according to $\mu$, then for any $0<\eta,\epsilon\leq\frac{1}{2}$, there exists an
\begin{equation*}
m=O\left(\frac{d+1}{\epsilon}\ln\frac{d+1}{\epsilon}+\frac{1}{\epsilon}\ln\frac{1}{\eta}\right)
\end{equation*}
such that
\begin{equation*}
\rho_{\T}\geq \varrho(\epsilon)
\end{equation*}
holds with probability at least $1-\eta$ when $n\geq m$. Note that $\varrho(\epsilon)\rightarrow1$ when $\epsilon\rightarrow0$.
\end{thm}
\begin{proof}
Theorem \ref{epsilon_net} shows that $\T$ is an $\epsilon$-net for range space $(D,B_{D})$ with probability at least $1-\eta$ when $n\geq m$. By lemma \ref{lem7}, we have
\begin{equation*}
\begin{split}
\rho_{\T}&=\frac{1}{\sqrt{d}}\int_{0}^{\sqrt{d}}\mu\left(D\cap\bigcup\limits_{x_{i}\in \T}B(x_{i},r)\right)dr \\
&\geq\frac{1}{\sqrt{d}}\int_{0}^{\sqrt{d}}\mu\left(\bigcup\limits_{A\in B_{D}^{\epsilon}(\frac{1}{2}r)}A\right)dr \\
&=\varrho(\epsilon).
\end{split}
\end{equation*}
\end{proof}
From this theorem, we know that a large number of samples lead to a sufficiently large $\rho_{\T}$ with a high probability.

In the previous sections, there is an assumption that every training point has only one single correct label, so we will naturally consider this special case in the sequel.

Denote
\begin{equation*}
D_{sin}=\{x\in D|\ tag(x)\in \{\{1\},\{2\},\cdots,\{K\}\}\ \},
\end{equation*}
\begin{equation*}
B_{D_{sin}}=\{B(x,r)\cap D_{sin} | x\in \R^{d}, r>0\},
\end{equation*}
\begin{equation*}
\mu_{sin}(A)=\frac{\mu(A\cap D_{sin})}{\mu(D_{sin})},
\end{equation*}
and $(D_{sin},B_{D_{sin}})$ is a range space with VC dimension at most $d+1$.
Let
\begin{equation*}
\T=\{x_{1},\cdots,x_{n}\}\subseteq D_{sin},
\end{equation*}
that is, the samples in $\T$ are drawn according to $\mu_{sin}$. As before, denote
\begin{equation*}
B_{D_{sin}}^{\epsilon}(r)=\{A\in B_{D}|A=B(x,s)\cap D, 0<s\leq r, \mu_{sin}(A)\geq\epsilon\},
\end{equation*}
\begin{equation*}
\varrho_{sin}(\epsilon)=\frac{1}{\sqrt{d}}\int_{0}^{\sqrt{d}}\mu\left(\bigcup\limits_{A\in B_{D_{sin}}^{\epsilon}(\frac{1}{2}r)}A\right)dr.
\end{equation*}

We have the following lemmas.
\begin{lemma}\label{lem9}
$\bigcup\limits_{A\in B_{D_{sin}}^{\epsilon}(\frac{1}{2}r)}A\subseteq D\cap\bigcup\limits_{x_{i}\in \T}B(x_{i},r)$ when $\T$ is an $\epsilon$-net for $(D_{sin},B_{D_{sin}})$.
\end{lemma}
\begin{lemma}\label{lem10}
$\lim\limits_{\epsilon\to0}\varrho_{sin}(\epsilon)= c_{sin}$, here
\begin{equation*}
c_{sin}=\frac{1}{\sqrt{d}}\int_{0}^{\sqrt{d}}\mu\left(\bigcup\limits_{A\in C_{D_{sin}}(\frac{1}{2}r)}A\right)dr\geq\mu(D_{sin}),
\end{equation*}
\begin{equation*}
C_{D_{sin}}(r)=\{A\in B_{D}|A=B(x,s)\cap D, 0<s\leq r, \mu_{sin}(A)>0\}.
\end{equation*}
\end{lemma}

From these two lemmas we deduce the following theorem.
\begin{thm}\label{thm4}
Let $\T=\{x_{1},x_{2},\dots,x_{n}\}$ be the training set drawn according to $\mu_{sin}$, then for any $0<\eta,\epsilon\leq\frac{1}{2}$, there exists an
\begin{equation*}
m=O\left(\frac{d+1}{\epsilon}\ln\frac{d+1}{\epsilon}+\frac{1}{\epsilon}\ln\frac{1}{\eta}\right)
\end{equation*}
such that
\begin{equation*}
\rho_{\T}\geq \varrho_{sin}(\epsilon)
\end{equation*}
holds with probability at least $1-\eta$ when $n\geq m$. Note that $\varrho_{sin}(\epsilon)\rightarrow c_{sin}$ when $\epsilon\rightarrow0$ for $c_{sin}\geq\mu(D_{sin})$.
\end{thm}

The proofs for Lemmas \ref{lem9}-\ref{lem10} and Theorem \ref{thm4} are very similar to those for Lemmas \ref{lem7}-\ref{lem8} and Theorem \ref{thm3}, respectively. We omit them here. It is noteworthy that $c_{sin}$ is intuitively very close to 1, even equal to 1. At worst, $c_{sin}$ is at least greater than $\mu(D_{sin})$ which may be quite large in practice, and the proof is similar to what we show in the Lemma in \ref{lem8}.

\section{Proof of Proposition \ref{pro:scale_independent}} \label{proof:scale_independent}
\begin{proof}
Let $\T$ be the training set and $\lambda$ be a positive constant greater than 1, $\widetilde{\T}=\T/\lambda$, $\widetilde{\mu}(A)=\mu(\lambda A)$, then
\begin{equation*}
\begin{split}
1-\rho_{\widetilde{\T}}&=\frac{1}{\sqrt{d}}\int_{0}^{\sqrt{d}}(1-h_{\widetilde{\T}}^{\widetilde{\mu}}(r))dr \\
&=\frac{1}{\sqrt{d}}\int_{0}^{\infty}(1-h_{\widetilde{\T}}^{\widetilde{\mu}}(r))dr \\
&=\frac{1}{\sqrt{d}}\int_{0}^{\infty}(1-h_{\T}^{\mu}(\lambda r))dr \\
&=\frac{1}{\lambda\sqrt{d}}\int_{0}^{\infty}(1-h_{\T}^{\mu}(r))dr \\
&=(1-\rho_{\T})/\lambda.
\end{split}
\end{equation*}
For the same reason,
\begin{equation*}
\begin{split}
CD(\widetilde{\T})&=\frac{1}{K(K-1)}\sum\limits_{i\neq j}(1-\rho(\widetilde{\T}_{i},\widetilde{\mu}_{j}))-\frac{1}{K}\sum\limits_{i}(1-\rho(\widetilde{\T}_{i},\widetilde{\mu}_{i})) \\
&=\frac{1}{\lambda K(K-1)}\sum\limits_{i\neq j}(1-\rho(\T_{i},\mu_{j}))-\frac{1}{\lambda K}\sum\limits_{i}(1-\rho(\T_{i},\mu_{i})) \\
&=CD(\T)/\lambda,
\end{split}
\end{equation*}
therefore
$$CC(\widetilde{\T})=CC(\T).$$
\end{proof}

\section{Proof of Proposition \ref{pro:empirical_bound}} \label{proof:empirical_bound}
\begin{proof}
Denote by $\widetilde{\T}_{c_1}=\{x_{i}\in \T\ |\ f$ is $c_{1}-$accurate at $x_i$$\}$, $\delta=\min(\delta_{0},\delta_{f}(c_{1}-c_{2}))$. For any $x_{i}\in \widetilde{\T}_{c_1}$, choose $k_{i}\in\{1,\cdots,K\}$ such that $tag(x_{i})=\{k_{i}\}$. From the definition of $\widetilde{\T}_{c_1}$ we know that $f_{k_{i}}(x_{i})>c_{1}$.

For any $x\in B(x_{i},\delta)\cap D$, from Proposition \ref{pro:separation_gap} we know that $tag(x)\cap tag(x_{i})\neq\varnothing$, and hence $k_{i}\in tag(x)$. On the other hand, $\|f(x)-f(x_{i})\|_{\infty}<c_{1}-c_{2}$, so we have $|f_{k_{i}}(x)-f_{k_{i}}(x_{i})|<c_{1}-c_{2}$, therefore
$$f_{k_{i}}(x)>f_{k_{i}}(x_{i})-(c_{1}-c_{2})>c_{1}-(c_{1}-c_{2})=c_{2},$$
which means that $f$ is $c_2-$accurate at $x$, that is to say
$$H_{c_{2}}^{f}\supseteq D\cap\bigcup\limits_{x_{i}\in \widetilde{\T}_{c_1}}B(x_{i},\delta).$$
Then
\begin{equation*}
\begin{split}
p_{c_{2}}(f)&=\mu(H_{c_{2}}^{f})\\
&\geq \mu\left(D\cap\bigcup\limits_{x_{i}\in \widetilde{\T}_{c_1}}B(x_{i},\delta)\right)\\
&=h_{\widetilde{\T}_{c_1}}^{\mu}(\delta)\\
&\geq 1-\frac{\sqrt{d}}{\delta}(1-\rho_{\widetilde{\T}_{c_1}})\\
&=1-\frac{\sqrt{d}}{\delta}(1-p_{c_{1}}^{\T}\rho_{\T}).
\end{split}
\end{equation*}
The second inequality can be derived from Proposition \ref{pro:h_estimation}.
\end{proof}

\section{Proof of Theorem \ref{thm:lower_bound}} \label{proof:lower_bound}
\begin{proof}
Define $\delta:=\min(\delta_{0},\delta_{f}(e^{-L_{f}^{max}}-c))$. Note that if $x_{i}\in \T$, and $x\in B(x_{i},\delta)\cap D$, then $\|x-x_{i}\|<\delta_{0}$, and hence $tag(x)\cap tag(x_{i})\neq \varnothing$, so that $k_{i}\in tag(x)$.

Because of
$$\|x-x_{i}\|<\delta_{f}(e^{-L_{f}^{max}}-c),$$
we have
$$\|f(x)-f(x_{i})\|_{\infty}<e^{-L_{f}^{max}}-c,$$
so that $|f_{k_{i}}(x)-f_{k_{i}}(x_{i})|<e^{-L_{f}^{max}}-c$. Therefore
$$f_{k_{i}}(x)>f_{k_{i}}(x_{i})-e^{-L_{f}^{max}}+c\geq e^{-L_{f}^{max}}-e^{-L_{f}^{max}}+c=c,$$
so that $f$ is c-accurate at $x$. Overall we obtain
\begin{equation*}
\begin{split}
p_{c}(f)&=\mu(H_{c}^{f}) \\
&\geq \mu\left(D\cap\bigcup\limits_{x_{i}\in \T} B(x_{i},\delta)\right)\\
&=h_{\T}^{\mu}(\delta)\\
&\geq 1-\frac{\sqrt{d}}{\delta}(1-\rho_{\T}).
\end{split}
\end{equation*}
The second inequality can be derived from Proposition \ref{pro:h_estimation}.
\end{proof}

\section{Proof of Theorem \ref{thm:deltaf_bound}} \label{proof:deltaf_bound}
\begin{proof}
We will prove it by contradiction. Consider two points $x_1$ and $x_2$ with different labels, and $\| x_1 - x_2 \| = \delta_{\T}$. 

If $\delta_f(e^{-L_f^{max}}-c) > \delta_{\T}/2$, then by the definition of $\delta_f(e^{-L_f^{max}}-c)$, $\|f(\frac{x_1+x_2}{2})-f(x_1)\|_{\infty} < e^{-L_f^{max}}-c$, since $\|\frac{x_1+x_2}{2} - x_1\| = \|\frac{x_2-x_1}{2} \|=\delta_{\T}/2$. Similarly, $\|f(\frac{x_1+x_2}{2})-f(x_2)\|_{\infty} < e^{-L_f^{max}}-c$. Then we have
$$\|f(x_1)-f(x_2)\|_{\infty} < 2(e^{-L_f^{max}}-c) \leq 2e^{-L_f^{max}}-1.$$
On the other hand, by the definition of $L_f^{max}$, $-\ln f_{k_1}(x_1) \leq L_f^{max}$ and $-\ln f_{k_2}(x_2) \leq L_f^{max}$, so $f_{k_1}(x_1) \geq e^{-L_f^{max}}$, $f_{k_2}(x_2) \geq e^{-L_f^{max}}$ and $f_{k_1}(x_2) \leq 1-f_{k_2}(x_2) \leq 1-e^{-L_f^{max}}$, therefore
\begin{equation*}
\begin{split}
\|f(x_1)-f(x_2)\|_{\infty} &\geq f_{k_1}(x_1)-f_{k_1}(x_2) \\
&\geq e^{-L_f^{max}} - (1-e^{-L_f^{max}}) \\
&= 2e^{-L_f^{max}} -1.
\end{split}
\end{equation*}
\end{proof}

\section{Proof of Theorem \ref{thm:bounds}}  \label{proof:bounds}
\begin{proof}
From Theorem \ref{thm:lower_bound} and assumption we know that
\begin{equation*}
\begin{split}
p(f)\geq p_{0.5}(f)&\geq1-\frac{\sqrt{d}}{\min(\delta_{0},\delta_{f}(e^{-L_{f}^{max}}-0.5))}(1-\rho_{\T}) \\
&\geq 1-\frac{\sqrt{d}}{\min(\delta_{0},\kappa\delta_{\T})}(1-\rho_{\T}) \\
&\geq 1-\frac{\sqrt{d}}{\kappa\delta_{0}}(1-\rho_{\T}), \\
\end{split}
\end{equation*}
which implies $\lim\limits_{\rho_{\T}\to 1}p(f)=1$. Note that $\kappa$ is less than $0.5$ and $\delta_{0}$ is only determined by the classification problem itself. The above inequality is easy to convert into form (\rnum 2).
\end{proof}

\section{Detailed information of data and parameters for training} \label{param_net}
First, we list the information concerning the data selection.
\begin{enumerate}
    \item MNIST: Last 55000 samples of the training set for training and all the 10000 samples of the test set for testing.
    \item CIFAR-10: First 49000 samples of the training set for training and all the 10000 samples of the test set for testing.
    \item CIFAR-100: First 49000 samples of the training set for training and all the 10000 samples of the test set for testing.
    \item COIL-20: 1200 samples whose end numbers of the figure names are not multiples of 6 for training and 240 samples whose end numbers are multiples of 6 for testing.
    \item COIL-100: 6000 samples whose end numbers of the figure names are not multiples of 30 for training and 1200 samples whose end numbers are multiples of 30 for testing.
    \item SVHN: First 50000 samples of the training set for training and first 10000 samples of the test set for testing.
\end{enumerate}
Parameters for networks are listed in Table~\ref{tab:param_net}.
\begin{table*}[htbp]
    \centering
    \begin{tabular}{cc|c|c|c|c|c|c|c|c|c|c|c|c|c}
        \toprule
        Data Set & Version & $\rnum{1}$ & $\rnum{2}$ & $\rnum{3}$ & $\rnum{4}$ & $\rnum{5}$ & $\rnum{6}$ & $\rnum{7}$ & $\rnum{8}$ & $\rnum{9}$ & $\rnum{10}$ & $\rnum{11}$ & $\rnum{12}$ & Best Error\\
        \midrule
        MNIST & Original & .02 & .02 & .05 & .02 & .02 & .04 & .02 & .02 & .04 & .01 & .02 & .03 & .01 \\
        CIFAR-10 & Original & .47 & .46 & .52 & .48 & .46 & .51 & .47 & .45 & .50 & .47 & .45 & .49 & .45 \\
        CIFAR-10 & Grey & .55 & .55 & .63 & .55 & .54 & .62 & .54 & .53 & .61 & .54 & .53 & .59 & .53 \\
        CIFAR-10 & Conv & .18 & .18 & .19 & .19 & .18 & .19 & .18 & .18 & .18 & .18 & .18 & .18 & .18\\
        SVHN & Original & .80 & .59 & .49 & .80 & .73 & .60 & .80 & .69 & .51 & .80 & .72 & .64 & .49 \\ 
        SVHN & Grey & .80 & .64 & .56 & .80 & .76 & .66 & .80 & .64 & .58 & .80 & .75 &.66 & .56 \\
        SVHN & Conv & .27 & .23 & .23 & .31 & .24 & .23 & .31 & .24 & .23 & .69 & .25 & .24 & .23 \\
        \midrule
        CIFAR-100 & Original(coarse) & .64 & .64 & .69 & .64 & .63 & .68 & .63 & .62 & .67 & .63 & .62 & .66 & .62 \\
        CIFAR-100 & Grey(coarse) & .74 & .73 & .79 & .74 & .73 & .78 & .74 & .72 & .78 & .74 & .72 & .77 & .72 \\
        CIFAR-100 & Conv(coarse) & .40 & .41 & .45 & .41 & .41 & .44 & .40 & .41 & .44 & .41 & .40 & .43 & .40 \\
        COIL-20 & Original & .08 & .05 & .05 & .07 & .06 & .05 & .08 & .05 & .05 & .07 & .05 & .03 & .03 \\
        \midrule
        CIFAR-100 & Original(fine) & .75 & .75 & .82 & .75 & .74 & .81 & .74 & .73 & .80 & .75 & .73 & .79 & .73 \\
        CIFAR-100 & Grey(fine) & .83 & .83 & .90 & .83 & .83 & .90 & .82 & .81 & .88 & .83 & .81 & .87 & .81 \\
        CIFAR-100 & Conv(fine) & .53 & .54 & .64 & .53 & .54 & .63 & .52 & .52 & .61 & .53 & .52 & .59 & .52 \\
        COIL-100 & Original & .02 & .01 & .01 & .03 & .02 & .01 & .03 & .02 & .01 & .02 & .02 & .01 & .01 \\
        \bottomrule
    \end{tabular}
    \caption{The results in this table were obtained as follows: We used the ReLU as the activation function and Adam as the optimizer. Furthermore, we chose several different network structures and learning rates. We then trained the networks for 10000 iterations with a batch size of 300, and selected the best test error as the final result for each training procedure. Each of the columns $\rnum{1}$-$\rnum{12}$ stands for a specific choice of network architecture and learning rate, as described in Table \ref{tab:param_detail}}
    \label{tab:param_net}
\end{table*}
\begin{table}[htbp]
    \centering
    \begin{tabular}{c|c|c|c}
        \toprule
        \diagbox{Structure}{Learning Rate} & $10^{-3}$ & $10^{-4}$ & $10^{-5}$ \\
        \midrule
        $[$Input 256 256 Output$]$ & $\rnum{1}$ & $\rnum{2}$ & $\rnum{3}$ \\
        \midrule
        $[$Input 256 256 256 Output$]$ & $\rnum{4}$ & $\rnum{5}$ & $\rnum{6}$ \\
        \midrule
        $[$Input 512 512 Output$]$ & $\rnum{7}$ & $\rnum{8}$ & $\rnum{9}$ \\
        \midrule
        $[$Input 512 512 512 Output$]$ & $\rnum{10}$ & $\rnum{11}$ & $\rnum{12}$ \\
        \bottomrule
    \end{tabular}
    \caption{Detailed setup for each case.}
    \label{tab:param_detail}
\end{table}

For generating convolution data, we choose the following structure
\begin{equation*}
\begin{split}
&conv[128]-relu-batchnorm-conv[256]-relu- \\
&batchnorm-pool-conv[512]-relu-batchnorm- \\
&conv[256]-relu-batchnorm-conv[64]-relu- \\
&batchnorm-pool-(extract\ data)-dense[512]- \\
&batchnorm-dropout-dense[128]-batchnorm- \\
&dropout-dense[output] \\
\end{split}
\end{equation*}
with kernel size $3\times 3$(strides $1$) and pool size $2\times 2$(strides $2$), then train this CNN with batch size $64$, learning rate $0.001$ and optimizer RMSProp for 5 epochs. After that, extract new data at location mentioned in above structure by feeding the data to the trained network.

\section*{References}
\bibliographystyle{elsarticle-harv} 
\bibliography{main}





\end{document}